%% file: main.tex
\documentclass[10pt,twocolumn,letterpaper]{article}

\usepackage{iccv}              %

\input{preamble}
\definecolor{iccvblue}{rgb}{0.21,0.49,0.74}
\usepackage[pagebackref,breaklinks,colorlinks,allcolors=iccvblue]{hyperref}

\def\paperID{1488} %
\def\confName{ICCV}
\def\confYear{2025}

\input{packages}

\input{macros}

\title{Generating Multi-Image Synthetic Data for Text-to-Image Customization}

\author{
Nupur Kumari$^{1}$ \qquad Xi Yin$^{2}$ \qquad Jun-Yan Zhu$^{1}$  \qquad Ishan Misra$^{2}$  \qquad Samaneh Azadi$^{2}$ \\ \\
$^{1}$Carnegie Mellon University \qquad
$^{2}$Meta
}

\begin{document}
\twocolumn[{%
\renewcommand\twocolumn[1][]{#1}%
\maketitle

\begin{center}
    \centering
    \includegraphics[width=0.95\linewidth]{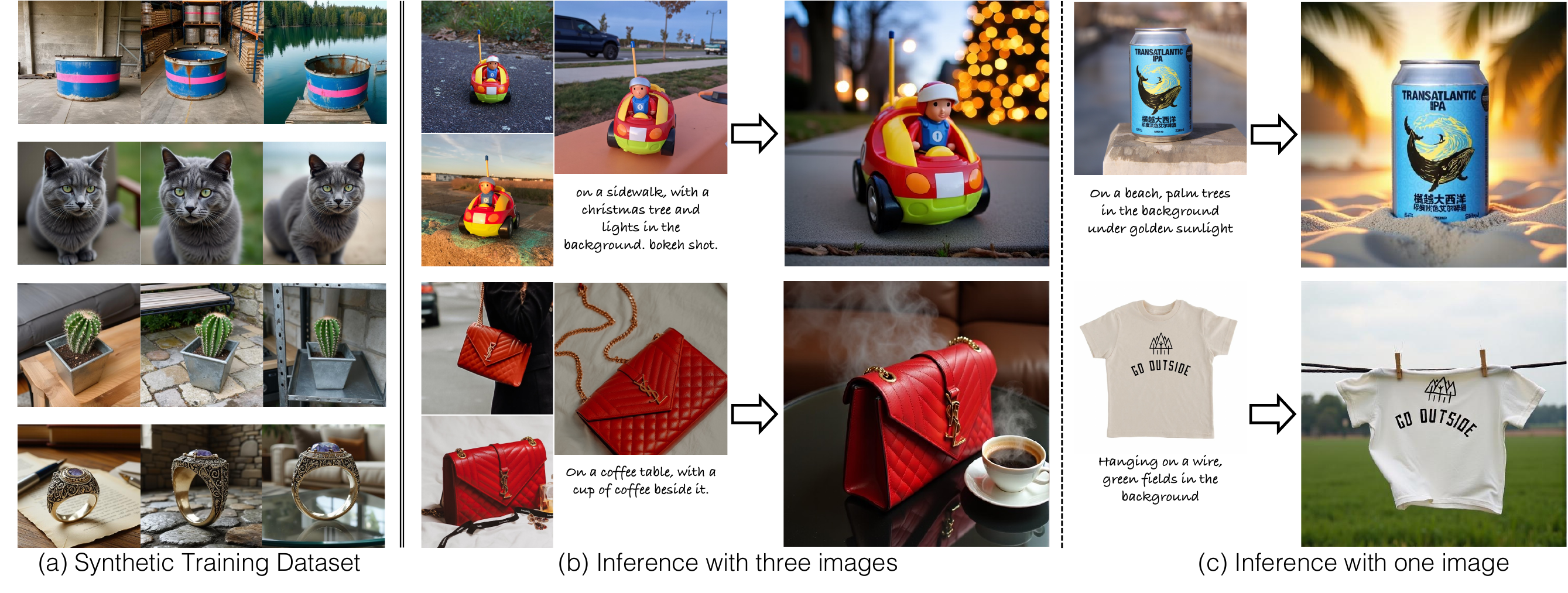}
    \vspace{-10pt}
\captionof{figure}{ (a) We propose a new pipeline for synthetic training data generation consisting of multiple images of the same object under different lighting, poses, and backgrounds. Given the dataset, we train a new encoder-based model customization method, which can take either (b) three or (c) one reference image of the object as input and successfully generate it in new compositions using text prompts.
}
\label{fig:teaser}
\end{center}

}]
 \maketitle
\input{sec/0_abstract}    
\input{sec/1_intro}

\input{sec/2_related_works}

\input{sec/3_method}

\input{sec/4_exp}

\input{sec/5_discussion}

\clearpage
\input{sec/7_ack}
{
    \small
    \bibliographystyle{ieeenat_fullname}
    \bibliography{main}
}
\appendix
\input{sec/6_appendix}

\end{document}

%% file: packages.tex
\usepackage{color,xcolor}
\usepackage{epsfig}
\usepackage{graphicx}
\usepackage{times}

\usepackage{comment}

\usepackage[accsupp]{axessibility} 
\usepackage{pifont}

\usepackage{microtype}
\usepackage[font=small]{caption}
\usepackage{arydshln}
\usepackage{tabularx}
\usepackage{adjustbox}
\usepackage{array}
\usepackage{booktabs}
\usepackage{colortbl}
\usepackage{float,wrapfig}
\usepackage{hhline}
\usepackage{multirow}
\usepackage{subcaption} %
\usepackage[percent]{overpic}

\usepackage[dvipsnames]{xcolor}

\usepackage{stmaryrd}
\usepackage{breqn}

\usepackage{amsmath,amsfonts,amssymb}
\usepackage{duckuments}
\usepackage{amsmath}

\usepackage{bm}
\usepackage{nicefrac}
\usepackage{microtype}
\usepackage{dsfont}
\usepackage{changepage}
\usepackage{extramarks}
\usepackage{fancyhdr}
\usepackage{setspace}
\usepackage{soul}
\usepackage{xspace}
\usepackage{hhline}
\usepackage{algorithmicx}
\usepackage{algorithm}
\usepackage{algpseudocode}
\usepackage{pifont}
\usepackage{amssymb}
\usepackage{tcolorbox}
\usepackage{makecell}
\usepackage{array}
\usepackage[pagebackref,breaklinks,colorlinks]{hyperref}
\usepackage{dashrule}
\usepackage{array}
\usepackage{enumitem}
\usepackage[title]{appendix}

%% file: macros.tex
\def\x{{\mathbf{x}}}
\def\v{{\mathbf{v}}}
\def\k{{\mathbf{k}}}
\def\q{{\mathbf{q}}}

\def\c{{\mathbf{c}}}

\makeatletter
\newcommand{\smallsym}[2]{#1{\mathpalette\make@small@sym{#2}}}
\newcommand{\make@small@sym}[2]{%
  \vcenter{\hbox{$\m@th\downgrade@style#1#2$}}%
}
\newcommand{\downgrade@style}[1]{%
  \ifx#1\displaystyle\scriptstyle\else
    \ifx#1\textstyle\scriptstyle\else
      \scriptscriptstyle
  \fi\fi
}
\makeatother

\newcommand*{\menlo}{\fontfamily{lmtt}\fontsize{9}{9}\selectfont }

\newcommand{\reffig}[1]{Figure~\ref{fig:#1}}
\newcommand{\refsec}[1]{Section~\ref{sec:#1}}
\newcommand{\refapp}[1]{Appendix~\ref{sec:#1}}
\newcommand{\reftbl}[1]{Table~\ref{tbl:#1}}

\newcommand{\refeq}[1]{Eqn.~\ref{eq:#1}}

\newcommand{\lblfig}[1]{\label{fig:#1}}
\newcommand{\lblsec}[1]{\label{sec:#1}}
\newcommand{\lbleq}[1]{\label{eq:#1}}

\newcommand{\ignorethis}[1]{}

\newcommand{\myparagraph}[1]{\vspace{1pt} \noindent \textbf{#1} \ }

\def\1{\bm{1}}

\newcolumntype{L}[1]{>{\raggedright\let\newline\\\arraybackslash\hspace{0pt}}m{#1}}
\newcolumntype{C}[1]{>{\centering\let\newline\\\arraybackslash\hspace{0pt}}m{#1}}
\newcolumntype{R}[1]{>{\raggedleft\let\newline\\\arraybackslash\hspace{0pt}}m{#1}}

\newcommand{\ignore}[1]{}

\renewcommand*{\thefootnote}{\arabic{footnote}}

\makeatletter
\DeclareRobustCommand\onedot{\futurelet\@let@token\@onedot}
\def\@onedot{\ifx\@let@token.\else.\null\fi\xspace}

\def\ie{i.e\onedot,\xspace}

\def\etal{\emph{et al}\onedot}
\makeatother

%% file: sec/0_abstract.tex
\begin{abstract}

Customization of text-to-image models enables users to insert new concepts or objects and generate them in unseen settings. Existing methods either rely on comparatively expensive test-time optimization or train encoders on single-image datasets without multi-image supervision, which can limit image quality. We propose a simple approach to address these challenges. We first leverage existing text-to-image models and 3D datasets to create a high-quality Synthetic Customization Dataset (\emph{SynCD}) consisting of multiple images of the same object in different lighting, backgrounds, and poses. Using this dataset, we train an encoder-based model that incorporates fine-grained visual details from reference images via a shared attention mechanism. Finally, we propose an inference technique that normalizes text and image guidance vectors to mitigate overexposure issues in sampled images. Through extensive experiments, we show that our encoder-based model, trained on \emph{SynCD}, and with the proposed inference algorithm, improves upon existing encoder-based methods on standard customization benchmarks. Please find the code and data at our \href{https://www.cs.cmu.edu/~syncd-project/} {website}.

\end{abstract}

%% file: sec/1_intro.tex
\section{Introduction}
\label{sec:intro}

Text-to-image models are capable of generating high-fidelity and realistic images given only a text prompt~\cite{peebles2023scalable,saharia2022photorealistic,rombach2022high,esser2024scaling}. Yet, text often falls short of describing rich visual details of real-world objects, such as the unique toy in \reffig{teaser}. What if the user wishes to generate images of this toy in new scenarios? This has given rise to the emerging field of model customization or personalization~\cite{ruiz2022dreambooth,gal2022image,kumari2023multi,chen2023subject,wei2023elite}, allowing us to generate new compositions of the object via text prompts, e.g., the toy on a sidewalk with a different background, as shown in \reffig{teaser}. Early \emph{optimization-based} works~\cite{gal2022image,dreamboothimpl,kumari2023multi} for the task require many fine-tuning steps on user-provided images of every new object --- a process both costly and slow. To address this, several \emph{encoder-based} methods~\cite{wei2023elite,chen2023subject,li2023blip,ye2023ip,song2024moma} learn an image encoder model with reference images as an additional conditional input. Thus, during inference, these methods can generate new compositions of the reference object in a single forward pass without expensive per-object optimization.

However, the lack of a dataset comprising multiple images of the same object in diverse poses, backgrounds, and lighting conditions has been a major bottleneck in developing these methods. Collecting such a large-scale \emph{multi-image} dataset from the internet is difficult, as real images are often not annotated with object identity. In this work, we aim to address this data shortage challenge using a new synthetic dataset generation method. This is challenging as we need to maintain the object's identity while generating multiple images with varying contexts. To achieve this, we leverage existing text-to-image models and 3D assets. Our first idea is to employ shared attention among foreground object regions while generating multiple images in parallel, ensuring visual consistency of the object across the images. Next, to ensure multi-view consistency for rigid objects, we use Objaverse~\cite{deitke2024objaverse} assets as a prior. Specifically, we use depth guidance and cross-view correspondence between different renderings to promote object consistency further. Finally, we filter out low-quality and inconsistent object images.

Given our Synthetic Customization Dataset, \emph{SynCD}, we train a new encoder-based model and propose an inference method for tuning-free customization.
Our encoder leverages cross-image shared attention to condition the output on fine-grained features of input reference images, improving object identity preservation. In summary, we introduce a pipeline to generate multiple images of an object in varying poses, lighting, and backgrounds using shared attention in text-to-image models and 3D assets. The encoder-based model trained with the Synthetic Customization Dataset, \emph{SynCD}, and with our proposed inference technique outperforms state-of-the-art encoder-based customization methods, including JeDi~\cite{zeng2024jedi}, Emu-2~\cite{Emu2}, and IP-Adapter~\cite{ye2023ip}.

%% file: sec/2_related_works.tex
\section{Related Works}
\myparagraph{Text-to-image models.}
With recent advancements in training methods~\cite{ho2020denoising,nichol2021improved,dhariwal2021diffusion,sauer2023stylegan,karras2022elucidating,karras2023analyzing,liu2022flow,yu2022scaling}, model architectures~\cite{peebles2023scalable,rombach2022high,esser2024scaling,ramesh2022hierarchical,kang2023scaling}, and datasets~\cite{schuhmann2021laion}, text-conditioned generative models have excelled at photorealistic generation while adhering to text prompts. Primarily among them are diffusion~\cite{lu2022dpm,rombach2022high} and flow~\cite{esser2024scaling,flux,liu2022flow,lipman2022flow} based models.
Their impressive generalization capability has enabled diverse  applications~\cite{parmar2023zero,richardson2023conceptlab,hertz2022prompt,mokady2023null,meng2021sdedit,hertz2024style,huang2024creativesynth,gu2024swapanything,ge2023expressive}. However, text as a modality can often be imprecise. This has given rise to various works on improving text alignment~\cite{chefer2023attend,ge2023expressive,liu2022compositional} and user control via additional image conditions~\cite{zhang2023adding,chen2022re}.

\myparagraph{Customizing text-to-image models.} A particular case of image-conditioned generation is the task of model customization or personalization~\cite{ruiz2022dreambooth,gal2022image,kumari2023multi}, which aims to precisely learn the concept shown in reference images, such as pets or personal objects, and compose it with the input text prompt. 
Early works in model customization fine-tune a subset of model parameters~\cite{kumari2023multi,han2023svdiff,hu2021lora,tewel2023key} or text token embeddings~\cite{gal2022image,voynov2023p+,zhang2023prospect,alaluf2023neural} on the few user-provided reference images with different regularization~\cite{ruiz2022dreambooth,kumari2023multi}. However, this fine-tuning process for every new concept is both time-consuming and computationally expensive. In contrast, our method focuses on training an encoder-based method without costly per-object optimization.

\begin{figure*}[!t]
    \centering
    \includegraphics[width=\linewidth]{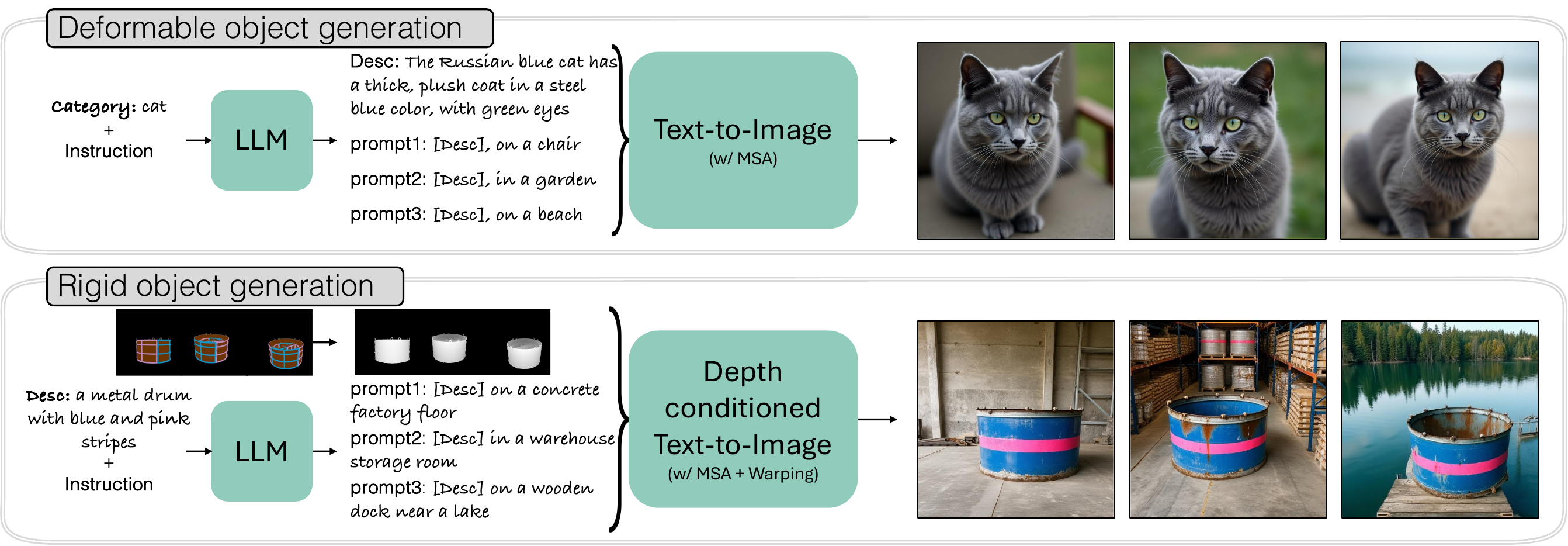}
    \caption{{\textbf{Dataset Generation Pipeline.} \textit{Top:} For deformable categories like cats, we use an object description combined with a set of background prompts, both suggested by an LLM, as input to generate multiple images of the same object in different contexts. \textit{Bottom:} For rigid objects, we use a depth-conditioned text-to-image model~\cite{zhang2023adding}. It takes depth map of Objaverse 3D assets~\cite{deitke2023objaverse} rendered from multiple views, its description~\cite{luo2024scalable}, and background context suggested by an LLM as input to generate the same object in varied poses and settings. We use Masked Shared Attention (MSA) and warping (in the case of rigid objects) to promote object consistency, as shown in \reffig{msa}. 
    }}
    \lblfig{dataset_generation}
    \vspace{-10pt}
\end{figure*}

\myparagraph{Encoder-based methods for customization}add additional image condition to the text-to-image model. To achieve this, many of the methods use pre-trained feature extractors to embed reference images into visual embeddings~\cite{li2023blip,song2024moma,wei2023elite,chen2024anydoor,xiao2024fastcomposer,parmar2025object}, which are then mapped to a text token embedding space. Some recent methods have also proposed learning a mapper between multimodal autoregressive models~\cite{touvron2023llama} and generative models to incorporate reference images as visual prompts~\cite{pan2023kosmos,Emu2}. Another commonly adopted design is the decoupled text and image cross-attention~\cite{ye2023ip,wei2023elite,ma2024subject}. Our training method is also motivated by this, but we insert fine-grained features via shared self-attention.

Most existing methods still rely on single-image or multi-view training datasets, with the same or limited background diversity. To prevent overfitting to the reference image pose or background, these are encoded in a compact feature space~\cite{li2023blip,song2024moma}, hurting identity preservation. To address this, we propose a new method for creating a synthetic dataset containing multiple images of the same object while having background and pose diversity. %
Our method is motivated by recent works in consistent character~\cite{tewel2024training,zhou2024storydiffusion} and multi-view generation~\cite{shi2023mvdream,deng2024flashtex,shi2023zero123++}, but tailored for the model customization task. JeDi~\cite{zeng2024jedi} and concurrent work OminiControl~\cite{tan2024ominicontrol} also created a synthetic dataset for customization. They use text prompting alone to generate images with the same objects. In contrast, our dataset curation method uses explicit constraints for object consistency guided by 3D assets, resulting in higher-quality training data.

%% file: sec/3_method.tex
\section{SynCD: Synthetic Customization Dataset}
Training encoder-based customization models requires a diverse dataset of different objects, each with multiple images in different contexts. 
To address the data shortage and collection challenge, we introduce a data curation pipeline for synthesizing diverse, high-quality image corpora. The pipeline consists of (1) Creating $N$ prompts per object, (2) Generating a \emph{set}  of $N$ images with a consistent object given the prompts, and (3) Dataset filtering to remove low-quality and inconsistent object images. For the dataset, we cover a large diversity of objects, which includes $75, 000$ rigid category assets from Objaverse~\cite{deitke2023objaverse} and $16$ deformable super-categories of animals, with approximately $100$ different subspecies. We explain each step of our pipeline in detail below.

\subsection{LLM assisted prompt generation}\lblsec{llm}
We design each prompt to have a detailed description of both the object and the background, as a detailed description of the object already helps enhance consistency. In the case of Objaverse, Cap3D~\cite{luo2024scalable} provides detailed captions for each asset, e.g., {\menlo a large metal drum with blue and pink stripes}. For deformable objects, we instruct the LLM to generate descriptive captions, e.g., {\menlo The Russian blue cat has a thick plush coat}. Based on the object description, we instruct the LLM~\cite{dubey2024llama} to then generate plausible background scene descriptions. 
Next, we combine one object description with multiple background descriptions and input it to the image generation step, as shown in \reffig{dataset_generation}. We use the Instruction-tuned LLama3~\cite{dubey2024llama} as the LLM and provide the instruction prompt we used in \refapp{details_dataset_gen}.

\subsection{Multi-image consistent-object generation} 
Given the prompts, we use the DiT~\cite{peebles2023scalable} -based FLUX model~\cite{flux} to generate images with a consistent object. FLUX consists of a series of MMDiT~\cite{esser2024scaling} blocks that gradually transform noise into a clean image in an encoded latent space~\cite{rombach2022high}. To enforce object consistency, we share the internal features across the images during the denoising process via a Masked Shared Attention (MSA) mechanism~\cite{tewel2024training,shi2023zero123++}. For rigid objects, we further leverage the depth and multi-view correspondence derived from Objaverse 3D assets.

\begin{figure}[!t]
    \centering
    \includegraphics[width=\linewidth]{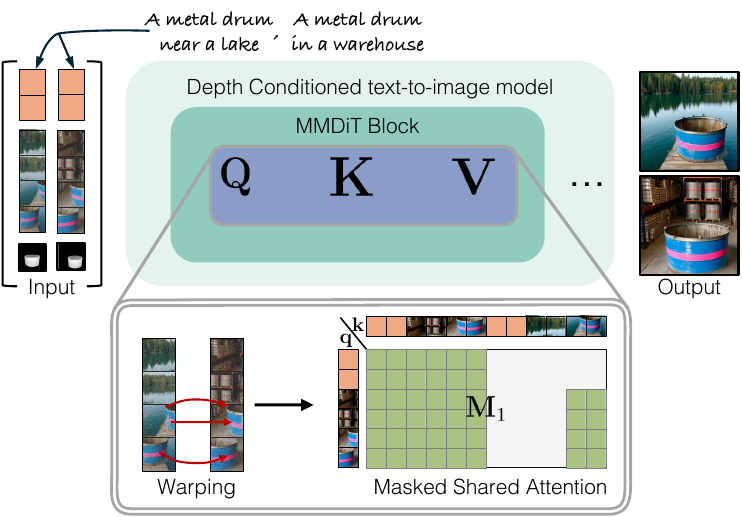}
    \caption{{\textbf{Feature warping and Masked Shared Attention (MSA) for object consistency.}  For rigid objects, we first warp corresponding features from the first image to the other. Then, each image feature attends to itself, and the foreground object features in other images. We show an example mask, $\mathbf{M}_1$, used to ensure this for the first image when generating two images with the same object. 
    }
    }
    \lblfig{msa}
    \vspace{-15pt}
\end{figure}

\myparagraph{Masked Shared Attention (MSA).} We modify the attention block of the diffusion model such that each image attends to itself as well as the foreground object regions of the other images. Thus, while generating $N$ images of an object in parallel with different prompts, in a particular attention layer, given query, key, and value features, $\q_i, \k_i, \v_i \in \mathbb{R}^{n \times d'} $, of the $i^{th}$ image, a shared attention performs the following operation:
\begin{equation}
    \begin{aligned}
    & \text{MSA}(\{\q_i, \k_i, \v_i\}_{i=1}^{N}) \equiv \\ &  \Big \{ \text{Softmax}\Big(\frac{\q_i [\k_1 \cdots \k_N]^T}{\sqrt{d'}} + \mathbf{M}_i \Big)[\v_1 \cdots \v_N] \Big \}_{i=1}^{N},\\
    \end{aligned}\lbleq{msa}
\end{equation}

where $d'$ is the feature dimension and $n$ is the sequence length of the image feature. Each $\q_i$ attends over the $N \times n$ features, and the `mask', \ie attention bias matrix $\mathbf{M}_i \in \mathbb{R}^{n \times (Nn)}$ ensures that the $i$-th image feature only attends to the object region of other images and ignores their background. Since a DiT model~\cite{peebles2023scalable} consists of joint text and image attention, the mask $\mathbf{M}_i$ is initialized so that text tokens of one image do not attend to other image tokens, as shown in \reffig{msa}. We also modify the Rotational Positional Embeddings (RoPe)~\cite{su2024roformer}, used in the DiT model, to be $NH \times W$ while generating the set of $N$ images.

MSA enables us to generate objects with similar visual features among all the images. However, it does not explicitly enforce 3D multi-view consistency, as qualitatively shown in \reffig{dataset_ablation2} in the Appendix. Therefore, for rigid objects with available 3D datasets like Objaverse, we use them to ensure multi-view consistency, as described next.

\myparagraph{Rigid object generation with MSA and 3D consistency.} Given an Objaverse asset, we render it from $N$ varying camera poses and feed the rendered depth map and captions generated in \refsec{llm} to a depth-conditioned FLUX model~\cite{fluxdepth}. During denoising process, Masked Shared Attention (MSA) is applied across all the images using the ground truth masks from the rendered depth map. Depth guidance ensures 3D shape consistency of the object across the images, while MSA encourages similar visual appearance. %
To further enhance multi-view consistency across generated images, we warp pair-wise corresponding features visible in the different views. For a representative example of generating two images with the same object, given latent features $f_i  \in \mathbb{R}^{(h \times w) \times d}$, $i \in \{1, 2\}$, the warping is calculated as:

\begin{equation}
    \begin{aligned}
   & \hat{f}_2(u,v) = \alpha f_1 (u + \Delta u, v + \Delta v) + (1 - \alpha) f_2(u,v),
\end{aligned}\lbleq{msa_objaverse}
\end{equation}
where for a given pixel $(u,v)$, $(u + \Delta u, v + \Delta v)$ denotes its corresponding location in the first image, $\alpha$ is a binary scalar, denoting if that location is visible in first image, and $f_1 (u + \Delta u, v + \Delta v)$ is the corresponding binliearly interpolated feature from the first image. \reffig{msa} shows an illustrative example. We apply warping for all pairs with appropriate masks and only during the early diffusion time steps. This increases multi-view consistency without introducing warping artifacts and allows flexibility for lighting variations.

\subsection{Dataset filtering}
Once generated, we filter out low-quality and inconsistent object images. We reject images with an aesthetic score~\cite{aesthetic} below  $6$. To measure object identity similarity, we use DINOv2~\cite{oquab2023dinov2} to remove images with an average pairwise feature similarity below 0.7 within their set. 
Our final dataset contains $\sim 95,000$ objects with $2$-$3$ images per object, uniformly distributed among rigid and deformable categories. \reffig{dataset_generation} shows our dataset generation pipeline.

\begin{figure}[!t]
    \centering
    \includegraphics[width=\linewidth]{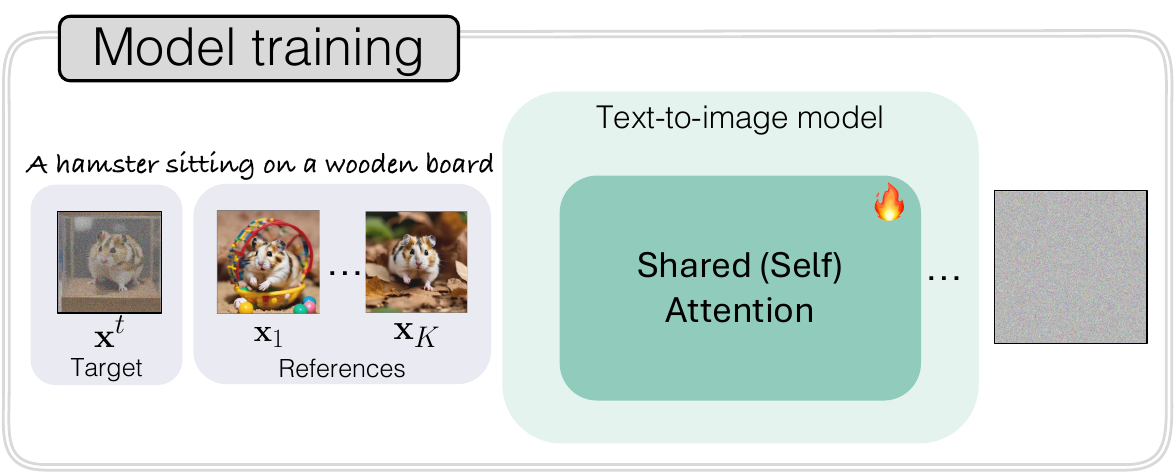}
    \vspace{-18pt}
    \caption{{\textbf{Training Method.} We condition the model on reference images, $\{\x_i\}_{i=1}^K$, using a Shared Attention mechanism, similar to \reffig{msa}. We extract fine-grained features of the reference images using the same model and have the target image features attend to the reference image features as well in the attention blocks.
    }
    }
    \lblfig{methoddiagram}
    \vspace{-15pt}
\end{figure}

\begin{figure*}[!t]
    \centering
    \includegraphics[width=0.95\linewidth]{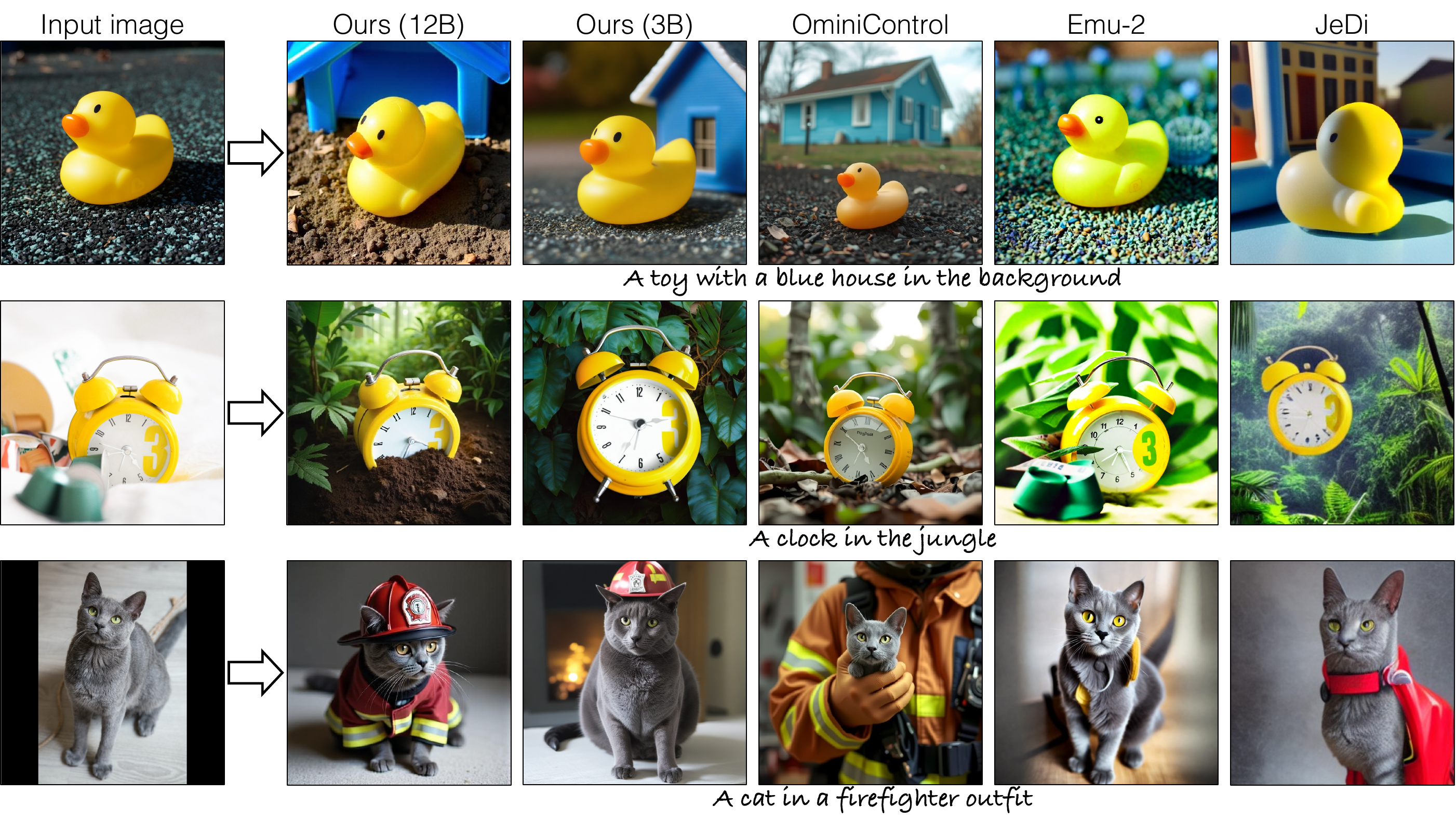}
    \vspace{-15pt}
    \caption{{\textbf{Results.} We compare our method qualitatively against other leading encoder-based baselines with a single reference image as input. We can successfully incorporate the text prompt while preserving the object identity similar to or higher than the baseline methods. We pick the best out of $4$ images for all methods. More qualitative samples are shown in \reffig{results_comparison_1ref_1} in the Appendix.
    }}
    \lblfig{results_comparison1}
    \vspace{-10pt}
\end{figure*}

\myparagraph{Discussion.} Our key insight for creating such a dataset is that synthesizing consistent object identities, using internal feature sharing and external 3D guidance, is far more scalable than collecting real-world data of multiple images with the same object. Moreover, generating such data is also more tractable than the task of model customization with real images, where access to the internal features and the object's true 3D geometry is not easily available.

\section{Our Method}
Given $K$ reference images $\{\x_i\}_{i=1}^K$ of an object, a customization method aims to learn $p(\x | \c, \{\x_i\}_{i=1}^K)$, i.e., the distribution of images aligned with both the input text prompt, $\c$, and object identity as shown in reference images. To achieve this, we fine-tune an existing text-to-image diffusion or flow-based model using our dataset. Given $N$ images of an object, we consider one of them as the target and the rest as references. For conditioning the generation on real reference images, we employ Shared Attention, similar to our dataset generation pipeline, as we explain below.

\myparagraph{Reference image conditioning.} 
During diffusion model training, the target image $\x$ is transformed to a noisy image $\x^{t} = \alpha^t \x + \sigma^t \epsilon$, $t\in [0, T]$, with $\x^T \sim \mathcal{N} (\mathbf{0}, \mathbf{I})$. The training objective is to denoise the input, $\x^{t}$, to $\mathbf{x}^{t-1}$, given the text prompt and reference images. To condition the denoising process on reference images, we concatenate their features with the target image features along the sequence dimension in each attention block of the diffusion model. The query features of the target image are subsequently updated by attending to both itself and the reference images' features. 

For the training loss, we adopt the velocity~\cite{salimans2022progressive} or flow prediction objective~\cite{lipman2022flow} for the diffusion and flow-based models, respectively, which is given as: 
\begin{equation}
    \begin{aligned}
        \mathbb{E}_{\x^t,t,\mathbf{c}, \epsilon \sim \mathcal{N} (\mathbf{0}, \mathbf{I})} ||\mathbf{v} - \mathbf{v}_{\theta} (\x^t, t, \mathbf{c}, \{\x_i\}_{i=1}^K) ||,
    \end{aligned}\lbleq{loss}
\end{equation}
where $\mathbf{v} \equiv \alpha^t\epsilon -\sigma^t\x$ and $\epsilon - \mathbf{x}$ for diffusion and flow models, parameterized by $\theta$, respectively, $t$ is the current timestep, $\alpha^t$ and $\sigma^t$ determine the noising ratio, and $\mathbf{v}_{\theta}$ is the predicted velocity/flow. The overall framework is as shown in \reffig{methoddiagram}.

\begin{figure*}[!t]
    \centering
    \includegraphics[width=0.98\linewidth]{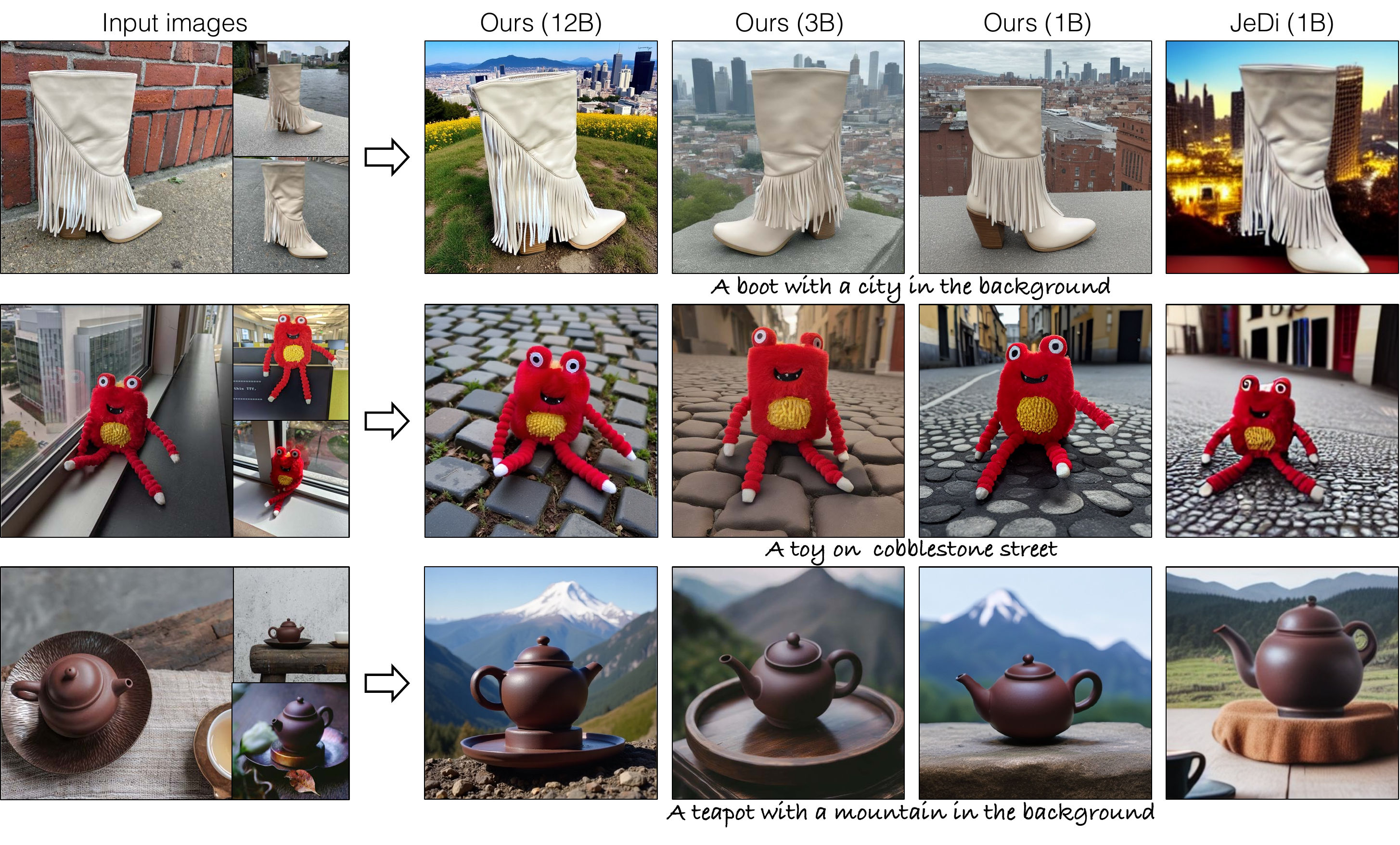}
    \vspace{-15pt}
    \caption{{\textbf{Results with 3 input images.} Here, we show qualitative samples of our method and JeDi~\cite{zeng2024jedi}, which can take multiple reference images as input. Though JeDi maintains high object identity alignment, the background and lighting can often be incoherent in the generated images. Comparatively, our method maintains higher image fidelity while following image and text conditions. We pick the best out of $4$ images for both methods. Zoom in for more details. We show more qualitative samples in \reffig{results_comparison_3ref} in the Appendix.
    }}
    \lblfig{results_comparison2}
    \vspace{-10pt}
\end{figure*}

\myparagraph{Inference} For final inference, we combine the classifier-free text and image guidance at every denoising step. %
However, directly combining them using previous work~\cite{brooks2023instructpix2pix} often leads to over-exposure issues in the generated image, especially at high image guidance, as shown in ~\reffig{sample_inf}. To mitigate this, we propose normalizing image and text guidance vectors. This helps us achieve better image alignment with the reference object while still following the text prompt. Our final inference is
\vspace{-8pt}
\begin{equation}
    \begin{aligned}
    & \epsilon_{\theta}(\x^t, \{\x_i\}_{i=1}^K, \varnothing) + \lambda_{I}\frac{||g||}{||g_{I}||} \cdot g_{I}  + \lambda_{\c}\frac{||g||}{||g_{c}||} \cdot g_{\c}, \\
 \text{where } & g_{I} = \epsilon_{\theta}(\x^t, \{\x_i\}_{i=1}^K, \varnothing)  - \epsilon_{\theta}(\x^t, \varnothing, \varnothing), \\
   & g_{\c} = \epsilon_{\theta}(\x^t, \{\x_i\}_{i=1}^K, \mathbf{c})  - \epsilon_{\theta}(\x^t, \{\x_i\}_{i=1}^K, \varnothing), \\
   & ||g|| = \min (||g_I||, ||g_{\c}||),
    \end{aligned}\lbleq{inference}
\end{equation}
where $t$ is the denoising timestep, $\epsilon_{\theta}$ is the model output, $g_{I}$ and $g_{c}$ are the image and text guidance vectors, and $\lambda_{I}$ and $\lambda_{\c}$ represent the guidance strength for the image and text. We scale the norm of the two guidance vectors to the minimum norm, allowing only $\lambda_{I}$ and $\lambda_{\c}$ to vary the relative strength of the image and text guidance. During inference, the number of reference images can vary from training since attention-based conditioning is agnostic to sequence length.

%% file: sec/4_exp.tex
\section{Experiments}

\myparagraph{Training details.} For a fair comparison with different baselines, we fine-tune a FLUX~\cite{flux} model, Ours (12B), and two Latent Diffusion Models (with 1B and 3B parameters) on our dataset with the same reference image conditioning. For the FLUX model, we only fine-tune attention layers with  LoRA~\cite{hu2021lora}. For the U-Net models, we initialize it with the IP-Adapter~\cite{ye2023ip} and fine-tune LoRA layers in the self-attention block and key-value projection matrices in the image cross-attention layers. More training and hyperparameter details are provided in \refapp{our_method_details}.

\myparagraph{Evaluation dataset.} Consistent with prior works~\cite{zeng2024jedi,song2024moma,pan2023kosmos}, we use DreamBooth~\cite{ruiz2022dreambooth} dataset consisting of $30$ objects with $4$-$5$ images each and 25 evaluation text prompts.

\myparagraph{Baselines.} We compare our method with leading encoder-based customization baselines, which include JeDi~\cite{zeng2024jedi}, IP-Adapter~\cite{ye2023ip,xfluxip}, Emu-2~\cite{Emu2}, Kosmos~\cite{pan2023kosmos}, BLIP-Diffusion~\cite{li2023blip}, and MoMA~\cite{song2024moma}. We also show a comparison with the concurrent work OminiControl~\cite{tan2024ominicontrol}. Sampling details for all are provided in \refapp{our_method_details}.

\myparagraph{Evaluation metric.} The goal of the text-conditional image customization task, given one or a few reference images, is to follow the input prompt while maintaining object identity and image fidelity. To measure the text alignment of generated images with the input prompt, we use CLIPScore~\cite{radford2021learning} and TIFA~\cite{hu2023tifa}. 
To evaluate the alignment of the object in generated images with the reference object, we compute similarity to reference images in DINOv2~\cite{oquab2023dinov2} feature space. Following recent works~\cite{zeng2024jedi,song2024moma}, we compute this similarity using a cropped and background-masked version of the image, denoted as MDINOv2-I, where the mask is computed by pre-trained object detectors~\cite{kirillov2023segment,zhou2022detecting,ren2024grounded}. Given the inherent tradeoff between text and image alignment metrics, we combine the two into a single metric, Geometric score~\cite{yan2023motion}, by taking the geometric mean of TIFA and MDINOv2-I. It is shown in~\cite{yan2023motion} that this geometric mean score is aligned better with the overall human preferences. 
In addition, we also conduct human evaluation to compare to prior works.

\begin{table*}[!t]
\centering
\setlength{\tabcolsep}{5pt}
\resizebox{0.75\linewidth}{!}{
\begin{tabular}{@{\extracolsep{4pt}}l c  cc c  c@{} }
\toprule
 \textbf{Method} & 
 \multicolumn{2}{@{} c}{\textbf{MDINOv2-I}$\uparrow$}
& \textbf{CLIPScore} $\uparrow$ & \textbf{TIFA} $\uparrow$ 
& \textbf{GeometricScore} $\uparrow$ \\
\cmidrule{2-3}
& \shortstack[c]{Background\\ change prompt} & \shortstack[c]{Property\\ change prompt}  & & \\
\midrule
 Kosmos~\cite{pan2023kosmos} &  0.636&0.638 & 0.287 &  0.729 & 0.679 \\
BLIP-Diffusion~\cite{li2023blip} &0.658&0.643    & 0.294 & 0.782 & 0.714 \\
MoMA~\cite{song2024moma}  & 0.616&0.620    & 0.320 & 0.867 & 0.730 \\
IP-Adapter~\cite{ye2023ip}  & 0.718 & 0.702 & 0.283 & 0.701 & 0.704\\
IP-Adapter Plus~\cite{ye2023ip}  &0.744& 0.737 & 0.270 & 0.615 & 0.675 \\
Emu-2~\cite{Emu2}   & 0.750& 0.736& 0.283 &  0.741  & 0.740  \\
JeDi~\cite{zeng2024jedi} & 0.771&0.775  & 0.292 &   0.789 & 0.780 \\
 Ours (1B)  & 0.806&0.773  & 0.303 &   0.830 & 0.801 \\
Ours (3B) & \textbf{0.822} &\textbf{0.789}  & 
\textbf{0.313} &  \textbf{0.863}  & \textbf{0.838}  \\
\hline 
 IP-Adapter (12B)~\cite{xfluxip} & 0.563 & 0.549 & 0.294 & \textbf{0.815} & 0.639\\
OminiControl (12B)~\cite{tan2024ominicontrol} & 0.650 & 0.527 & 0.302 & 0.808  & 0.685 \\
Ours (12B) & \textbf{0.778} & \textbf{0.771} & \textbf{0.306} & 0.786 & \textbf{0.780} \\
\bottomrule
\end{tabular}
}
\vspace{-8pt}
\caption{ \textbf{Quantitative comparison}. We compare our method against other encoder-based methods on image alignment and text alignment metrics. Our method performs better or on par with other baselines on the combined GeometricScore metric, even when compared across different model scales. For reference, the all-pairwise MDINOv2 similarity between reference images themselves is $0.851$. }
\label{tbl:metrics}
\vspace{-8pt}
\end{table*}

\subsection{Comparison to Prior Works}
\subsubsection{Qualitative Comparison}
We show sample comparisons of our method against other encoder-based methods in \reffig{results_comparison1} and \ref{fig:results_comparison2}. Our method more effectively incorporates the text prompt while keeping the object identity and image fidelity, e.g., the cat in the firefighter outfit in $3^{\text{rd}}$ row of \reffig{results_comparison1}. In contrast, baseline methods can struggle with incorporating the text prompt or have low object identity preservation. With $3$ reference images as input in \reffig{results_comparison2}, although JeDi~\cite{zeng2024jedi} achieves high identity preservation, it can result in reduced image quality, with inconsistency in lighting and background scene.

\subsubsection{Quantitative Comparison}
\myparagraph{Automatic scores.} \reftbl{metrics} compares our method with encoder-based baselines. We measure the MDINOv2-I metric on two subsets: prompts that only change the background and those that modify object appearance, e.g., {\menlo cube-shaped or wearing sunglasses}, with the latter expected to yield lower image similarity in comparison. \reftbl{metrics} shows our method's performance with $3$ input reference images. All variants of our model perform better or on par with the baselines in the overall Geometric Score, last column in~\reftbl{metrics}. While Ours (3B) achieves a higher DINOv2-I score, Ours (12B) generates better fidelity images (\reffig{results_comparison2}) with increased viewpoint and background diversity, as shown in \reffig{sdxl_vs_flux} in the Appendix. Our method also works with $1$ reference image as shown in \reffig{results_comparison1}. We report quantitative metrics with one input image in \reftbl{tuningbased_metrics} in the Appendix, which also shows a comparison with optimization-based approaches, with our method having competitive image alignment and better text alignment.

Though quantitative metrics measure image and text alignment, they can struggle with capturing overall quality and favor methods that copy-paste the target object on a new background. Thus, for a more comprehensive evaluation, we conduct a pairwise human study next.

\myparagraph{Human evaluation.}
In each study, participants view two generated images (from our method and a baseline) alongside the text prompt and $3$ reference images. We ask them to select the preferred image based on three criteria: (1) Consistency with the reference object (image alignment), (2) Alignment with the text prompt (text alignment), and (3) Overall quality and photorealism (quality). They also indicate the specific criterion or criteria for their selection. \reftbl{human_eval} shows the results compared to the three competing methods from \reftbl{metrics}, i.e., Emu-2~\cite{Emu2}, JeDi~\cite{zeng2024jedi}, and OminiControl~\cite{tan2024ominicontrol}. Our method is preferred over the baselines according to all evaluation criteria, showing the effectiveness of our synthetic dataset. To ensure valid responses, participants complete a practice test, and only those with correct responses are considered. We gather more than $300$ valid responses per comparison and provide further details regarding the study in \refapp{appendix_eval_details}.

\begin{table}[!t]
\centering
\setlength{\tabcolsep}{5pt}
\resizebox{\linewidth}{!}{
\begin{tabular}{l cccc  }
\toprule
\textbf{Method} & \multicolumn{4}{@{} c}{\textbf{Human preference} (in $\%$)$\uparrow$}
\\
 & \shortstack[c]{Text\\alignment} &  \shortstack[c]{Image\\alignment} &   \shortstack[c]{Photo-\\realism}     &   \shortstack[c]{Overall\\preference} \\
\midrule

Ours (1B) vs JeDi & 69.51 & 63.05  & 80.89 & 68.19 \\
Ours (3B) vs Emu-2  & 70.49 & 66.88  & 64.66 & 66.74 \\
Ours (12B) vs OminiControl & 56.27 & 58.30 & 54.47 &  58.02 \\

\bottomrule
\end{tabular}
}
\vspace{-8pt}
\caption{ \textbf{Human preference.} Here, we compare the pairwise preference of our method against the competing methods from \reftbl{metrics}, i.e., Emu-2~\cite{Emu2}, JeDi~\cite{zeng2024jedi}, and OminiControl~\cite{tan2024ominicontrol}, while keeping the same model scale. The standard error for all is within $\pm 5\%$. %
}

\label{tbl:human_eval}
\vspace{-12pt}
\end{table}

\subsection{Ablation Study}\lblsec{model_ablation}
In this section, we conduct various ablations regarding different components of our method.

\myparagraph{Model training.} 
To show the effectiveness of training the customization model with shared attention, we fine-tuned the baseline IP-Adapter Plus model, a similar scale model as Ours (3B), on our dataset. As shown in \reftbl{ablation} (row 3), simply fine-tuning with our dataset already improves its performance while using similar models and inference protocols, which highlights the contribution of our dataset. Subsequently, adding the reference condition via shared attention further boosts performance, \reftbl{ablation} (row 4), showing its effectiveness. It also allows the use of multiple reference images during inference, \reftbl{ablation} (row 5), improving performance as we increase the number of reference images. We show qualitative samples in \reffig{model_ablation_fig} in the Appendix.

\begin{table}[!t]
\centering
\setlength{\tabcolsep}{5pt}
\resizebox{\linewidth}{!}{
\begin{tabular}{ll cc c c}
\toprule
\multirow{3}{*} & \textbf{Method}
& \multicolumn{2}{@{} c}{\textbf{MDINOv2-I}$\uparrow$} 
& \textbf{TIFA} $\uparrow$ 
&  \textbf{Geometric} $\uparrow$ \\
& &  Background &  Property & & \textbf{Score} \\
& & change prompt & change prompt & & \\
\midrule
\multirow{4}{*}{1-input} 
& IPAdapter Plus  & 0.744 & 0.737 & 0.615 & 0.675 \\
& + our inference & 0.719 & 0.668 &  0.816 & 0.756 \\ 
& + SynCD & 0.766  & 0.695   & 0.901 & 0.819 \\
& + MSA (Ours-3B) & 0.777 & 0.708 & 0.902 &  0.825 \\
\multirow{1}{*}{3-input} & + MSA (Ours-3B) & \textbf{0.822} & \textbf{0.789} & \textbf{0.863}  & \textbf{0.838}  \\
\bottomrule
\end{tabular}
}
\vspace{-8pt}
\caption{ \textbf{Model ablation}. We add different components of our method- modified inference, our dataset, and shared attention to the baseline IP-Adapter Plus method - and show a gradual increase in performance. MSA also enables the effective use of multiple reference images as input, thus significantly helping with image alignment as we increase the number of reference images to three.
}

\label{tbl:ablation}
\vspace{-10pt}
\end{table}

\begin{figure}[t]
    \centering
    \includegraphics[width=\linewidth]{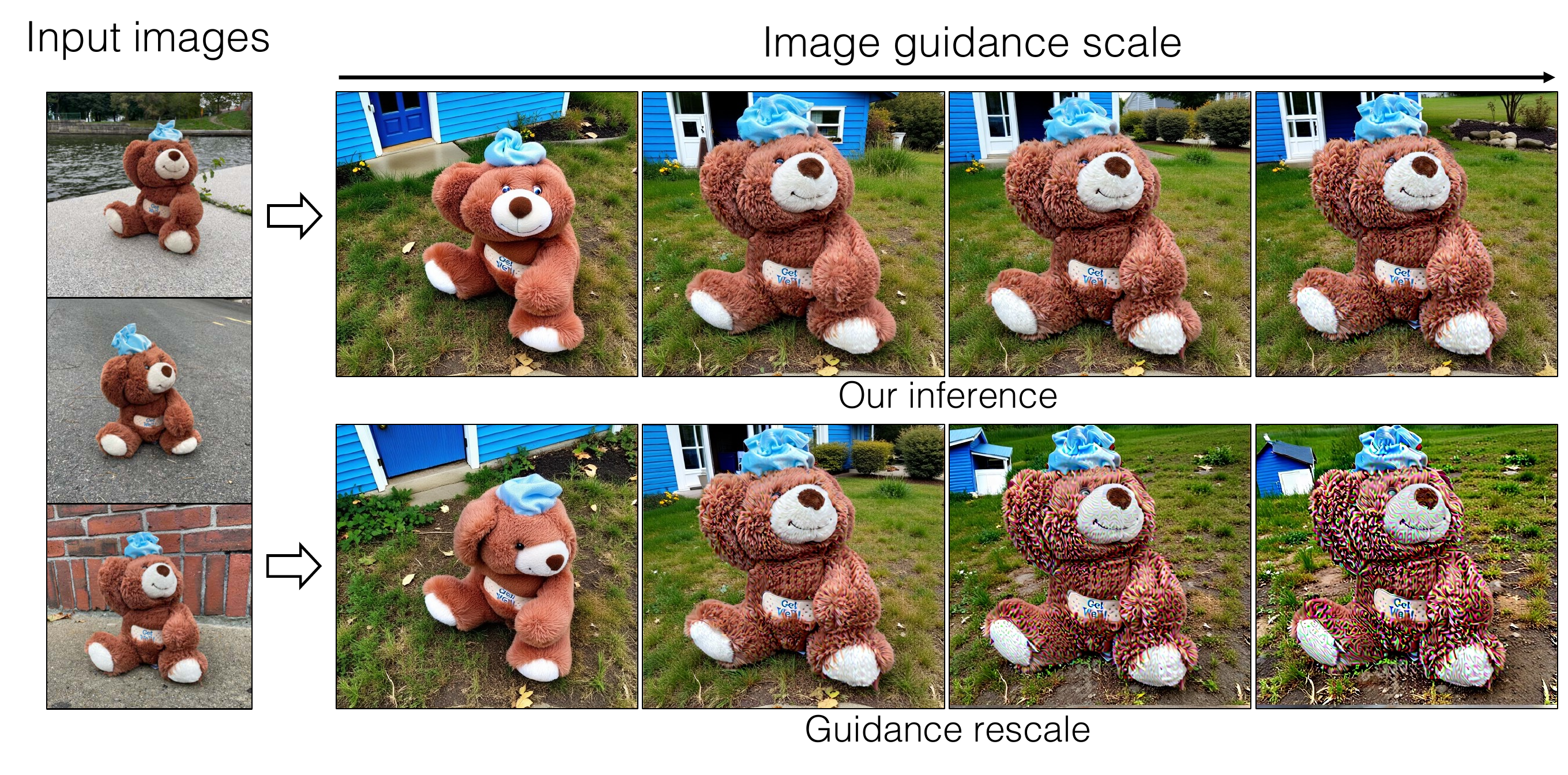}
    \vspace{-25pt}
    \caption{{\textbf{Our inference} comparison with guidance rescale~\cite{lin2024common} using Ours (12B) model. The caption is {\menlo A stuffed animal with a blue house in the background}. As image guidance is increased from $1$ to $5$, our inference follows the text prompt while increasing the image similarity without artifacts, thus allowing us to use higher image guidance in general. Please zoom in for details
    }}
    \lblfig{sample_inf}
    \vspace{-15pt}
\end{figure}

\myparagraph{Modified guidance inference.} 
Here, we compare our inference approach (\refeq{inference}) to guidance rescale~\cite{lin2024common}. Guidance rescale was also proposed to mitigate image saturation, but in the vanilla text-to-image generation pipeline. As \reffig{sample_inf} shows, increasing the guidance strength in our method preserves image fidelity while incorporating the text and image conditions. We also evaluate the baseline IP-Adapter Plus~\cite{ye2023ip} with our modified inference. This improves its TIFA score from $0.615$ to $0.816$, with only a minor decrease in image alignment, as shown in \reftbl{ablation} (row 2). We show more analysis and qualitative samples in \refapp{appendix_ablation}.

\begin{figure}[t]
    \centering
    \includegraphics[width=\linewidth]{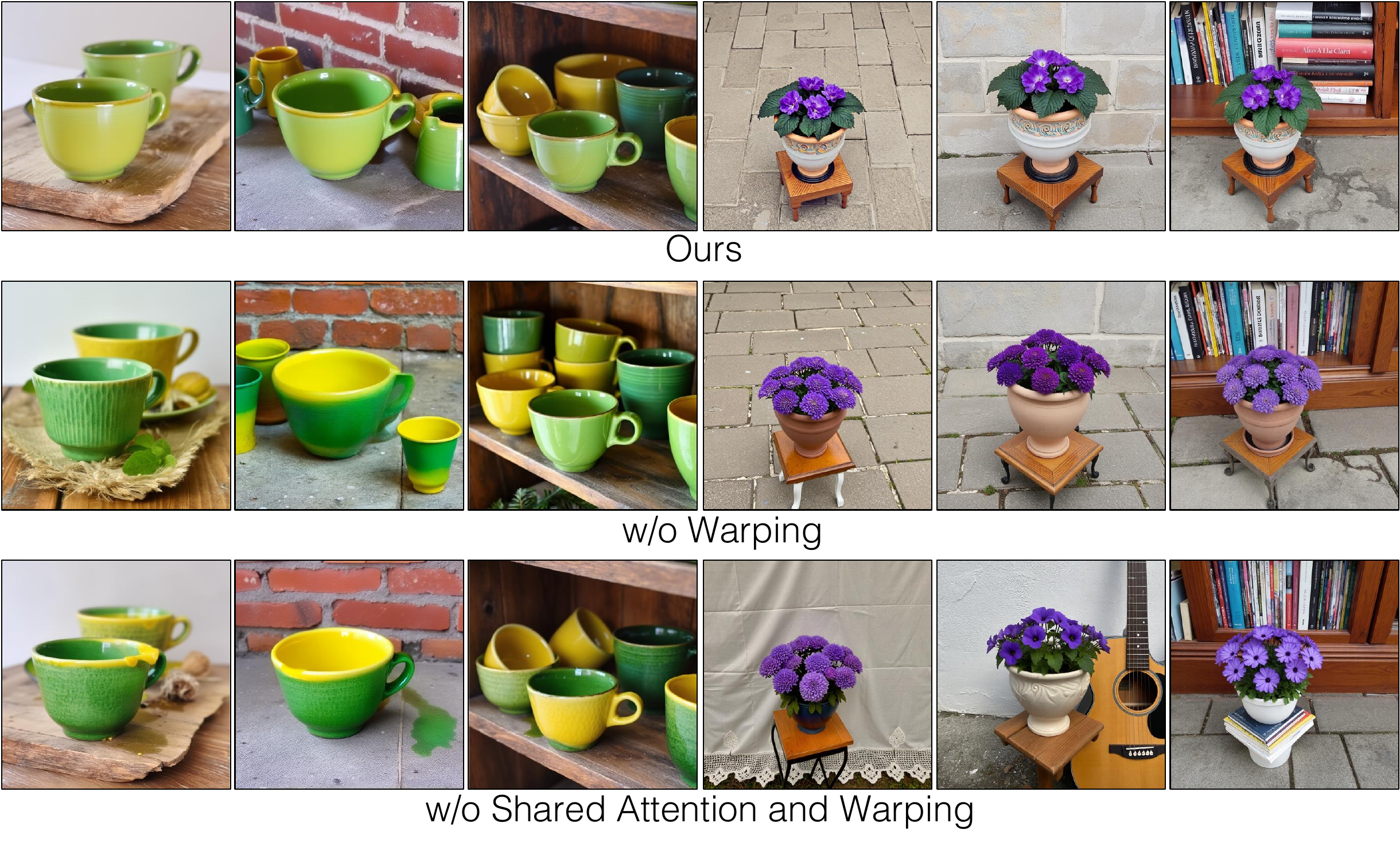}
     \vspace{-25pt}
    \caption{{\textbf{Dataset generation ablation.} \textit{Top:} our synthetic training images. \textit{Middle:} removing warping reduces multi-view consistency, e.g., the colors of the center cup in the left column or the flower pot in the right column. \textit{Bottom:} removing both warping and MSA
    further hurts visual consistency. Zoom in for details.
    }}
    \vspace{-8pt}
    \lblfig{sample_images_2}
\end{figure}

\begin{table}[!t]
\centering
\setlength{\tabcolsep}{5pt}
\resizebox{0.8\linewidth}{!}{
\begin{tabular}{l lc}
\toprule
& \textbf{Method} & \textbf{DINOv2-I}$\uparrow$ 
\\
\midrule
\multirow{3}{*}{\shortstack[l]{Rigid\\ categories}} & Ours & \textbf{0.598} 
\\
& w/o Warping &  0.572
\\
& w/o MSA and Warping  &  0.495 
\\
\hline
\multirow{3}{*}{\shortstack[l]{Deformable\\ categories} } & Ours & \textbf{0.700} 
\\
& w/o MSA & 0.626 
\\
& w/o Detailed description & 0.564 
\\
\bottomrule
\end{tabular}
}
\vspace{-8pt}
\caption{ \textbf{Dataset curation ablation.} MSA consistently enhances intra-cluster DINOv2-I similarity. Additionally, warping, in the case of rigid objects, further improves it. The qualitative benefits of both are shown in \reffig{sample_images_2}. 
}
\label{tbl:dataset_ablation}
\vspace{-15pt}
\end{table}

\myparagraph{Dataset curation.} We ablate different steps of the dataset generation to analyze their respective contributions. We compute the average intra-cluster similarity using DINOv2 features, where a cluster comprises the $N$ images generated in parallel with the same object. \reftbl{dataset_ablation} shows that MSA consistently improves intra-cluster similarity, and for rigid object generation using Objaverse assets, feature warping further enhances it. We find it specifically beneficial in promoting cross-view consistency between the object in the images, e.g., the consistent cup colors in $1{\text{st}}$ row (left column) of \reffig{sample_images_2}. For deformable objects, providing descriptive prompts in addition to MSA proves crucial.

\myparagraph{Role of dataset size on model performance.} Here, we examine how dataset size affects model performance. \reftbl{dataset_ablation_size} shows image-alignment metrics for models trained on progressively larger datasets. Since Ours (3B) model is initialized from IP-Adapter Plus~\cite{ye2023ip}, it has high image alignment, even when fine-tuned on just $100$ samples, but exhibits lower pose and background diversity overall, as shown in \reffig{sdxl_vs_flux} in the Appendix. We find that the dataset size is more crucial for the larger Ours (12B) model, yielding greater improvements in image alignment with increasing dataset size. \reffig{dataset_size} shows the qualitative samples, which indicate that training on increasingly larger datasets enhances background and pose diversity while capturing fine object details. Comparatively, models trained on fewer samples suffer from overfitting and reduced diversity. Additionally, with a fixed dataset size, greater category diversity improves performance, as our analysis in \refapp{appendix_ablation} shows. 

In the Appendix, we also include results on CustomConcept101 evaluation benchmark~\cite{kumari2023multi}, a comparison with using the OminiControl synthetic dataset~\cite{tan2024ominicontrol} for training, and more qualitative samples.

\begin{figure}[t]
    \centering
    \includegraphics[width=\linewidth]{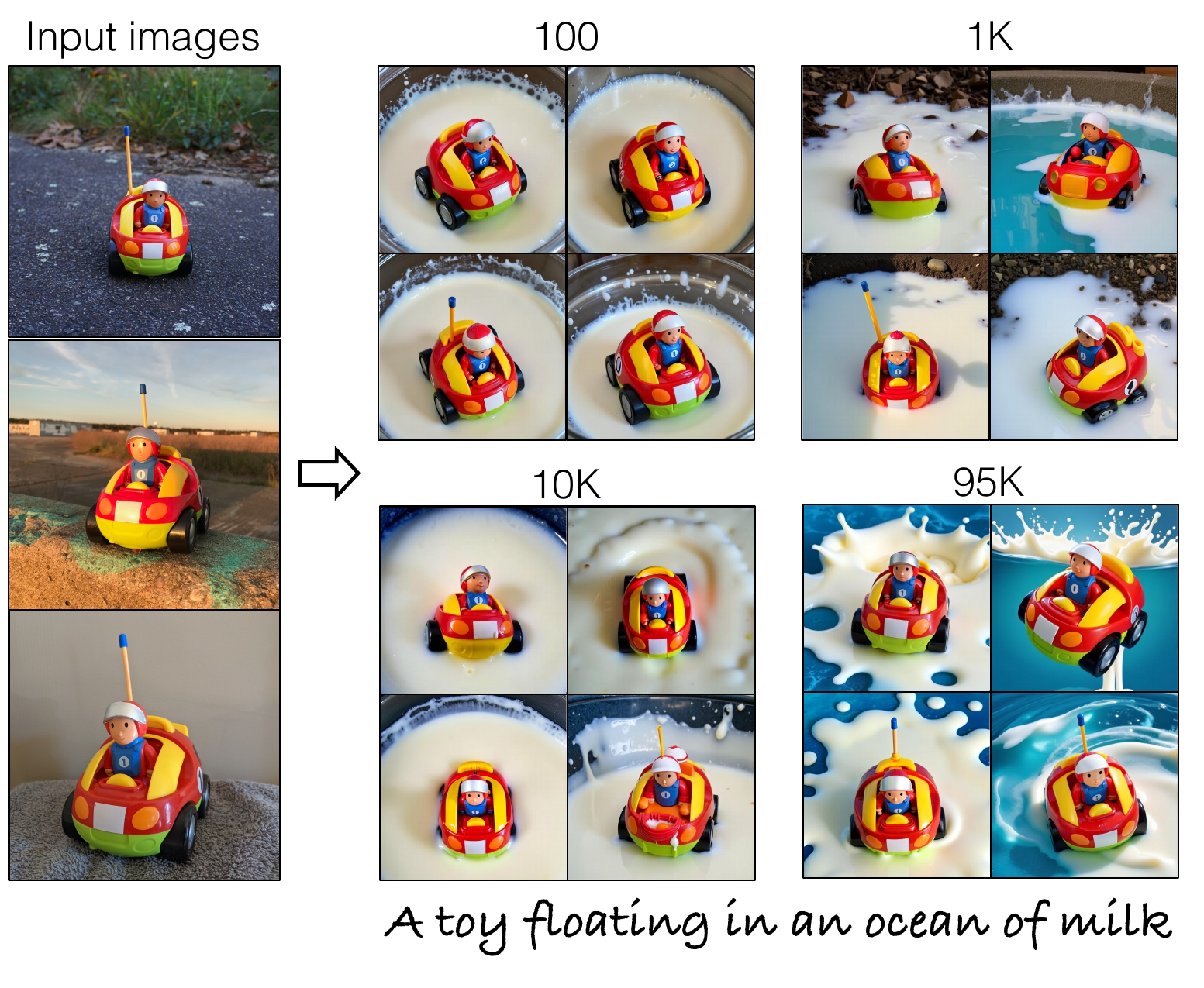}
    \vspace{-25pt}
    \caption{{\textbf{Dataset size.} Increasing training samples from $100$ to $1K$, $10K$, and $95K$ yields improvements in object identity preservation while generating diverse backgrounds and viewpoints without overfitting issues. Please zoom in for more details.
    }}
    \lblfig{dataset_size} 
\end{figure}

\begin{table}[!t]
\centering
\resizebox{0.7\linewidth}{!}{
\begin{tabular}{l cccc}
\toprule
\textbf{Method} & \multicolumn{4}{@{} c}{\textbf{MDINOv2-I}$\uparrow$}
\\
\cmidrule{2-5}
 & 100 & 1K & 10K & Ours(95K)  \\
\midrule
\textbf{Ours (3B)} & 0.790 & 0.805 &  0.810 & 0.813 \\
\textbf{Ours (12B)} & 0.736  & 0.762  & 0.763  & 0.774   \\
\bottomrule
\end{tabular}
}
\vspace{-8pt}
\caption{ \textbf{Dataset size vs. performance.} With the increase in dataset size, performance increases, both in terms of image alignment as shown here and overall photorealism and background diversity as shown in \reffig{dataset_size}, specifically for larger models like Ours (12B).
}\label{tbl:dataset_ablation_size}
\vspace{-12pt}
\end{table}

%% file: sec/5_discussion.tex
\section{Discussion and Limitations}
In this work, we focus on encoder-based model customization and propose advancements to address current limitations. To overcome the lack of training data, we have created a synthetic dataset by generating multiple images with consistent objects using Masked Shared Attention and 3D asset priors. We have also proposed an improved model architecture and inference technique. Our approach outperforms existing encoder-based methods while being on par with existing computationally expensive optimization-based approaches. %

Though promising, our dataset has room for improvement. First, our work focuses on single-object images. Extending our method to multi-object, multi-view datasets would be a meaningful next step. Second, incorporating recent advances in text-to-3D and video generative models, along with scaling dataset generation to include a wider range of objects, could further enhance its quality and diversity.

%% file: sec/7_ack.tex
\myparagraph{Acknowledgment.}
We thank Kangle Deng, Gaurav Parmar, and Maxwell Jones for their helpful comments and discussion and Ruihan Gao and Ava Pun for proofreading the draft. This work was partly done by Nupur Kumari during the Meta internship. The project was partly supported by the Packard
Fellowship, the IITP grant funded by the Korean Government (MSIT) (No. RS-2024-00457882, National AI Research Lab Project), NSF
IIS-2239076, and NSF ISS-2403303.

%% file: sec/6_appendix.tex
\renewcommand{\thefootnote}{\arabic{footnote}}
\clearpage
\noindent{\Large\bf Appendix}
\vspace{5pt}

In \refsec{appendix_result} and \ref{sec:appendix_ablation}, we show more qualitative samples of our method, its comparison to the baselines, and more ablation studies. Then, in \refsec{details}, we provide implementation details related to our dataset generation, model training, and inference. Finally, in \refsec{limitation}, we discuss our work's limitations and societal impact.

\section{Additional Comparison with Prior Works}\lblsec{appendix_result}

\myparagraph{CustomConcept101 Benchmark.}
Though DreamBooth~\cite{ruiz2022dreambooth} is a widely used evaluation dataset, CustomConcept101~\cite{kumari2023multi} is more diverse with $101$ unique concepts. Here, we also compare our model (3B) with open-source baseline models of similar scale, i.e., Emu-2~\cite{Emu2} and IP-Adapter~\cite{ye2023ip}, on this dataset. As shown in \reftbl{customconcept101}, our method performs better in identity preservation compared to both baselines while also yielding higher text alignment as indicated by CLIPScore~\cite{radford2021learning} and TIFA~\cite{hu2023tifa} metrics. 

\myparagraph{Qualitative Comparison.}
In \reffig{sdxl_vs_flux}, we compare Ours (12B) and Ours (3B) qualitatively and show that Ours (12B), fine-tuned from FLUX~\cite{flux}, has better generalization with greater viewpoint and background diversity in generated samples. This also explains a relatively lower image alignment according to DINOv2-I metrics for Ours (12B) than Ours (3B) in Table 1 of the main paper. Figure~\ref{fig:results_comparison_3ref} and \ref{fig:results_comparison_1ref_1} show more visual comparison of our method against the baselines on the DreamBooth~\cite{ruiz2022dreambooth} dataset with three and one reference images as input, respectively. More random samples of our method on DreamBooth and CustomConcept101 evaluation benchmarks are shown in \reffig{random_samples_1} and \ref{fig:random_samples_2}.

\myparagraph{Comparison to optimization-based methods}
We also compare our method against optimization-based approaches in Table~\ref{tbl:tuningbased_metrics}. For a single input image, we compare it with Break-a-Scene~\cite{avrahami2023break}, which also uses one image. With $3$ reference images, we benchmark against LoRA~\cite{loraimplementation,dreamboothimpl}. As shown in \reftbl{tuningbased_metrics}, our method achieves comparable performance in image alignment while improving text alignment, suggesting reduced overfitting to the reference images. In addition, LoRA fine-tuning with FLUX~\cite{flux} takes $10$ minutes at $512$ resolution. Whereas, once trained, our model can generate a sample for any new object in a feed-forward manner in $15$ and $25$ seconds using $1$ or $3$ reference images, at the same resolution, respectively (all times measured on an H100 GPU). \reffig{tuning_based_samples1} shows sample comparisons of our method with optimization-based approaches Break-a-Scene~\cite{avrahami2023break} and LoRA~\cite{hu2021lora,loraimplementation}. When compared to them, our method performs on par in identity preservation while better following the text prompt.

\begin{figure}[t]
    \centering
    \includegraphics[width=\linewidth]{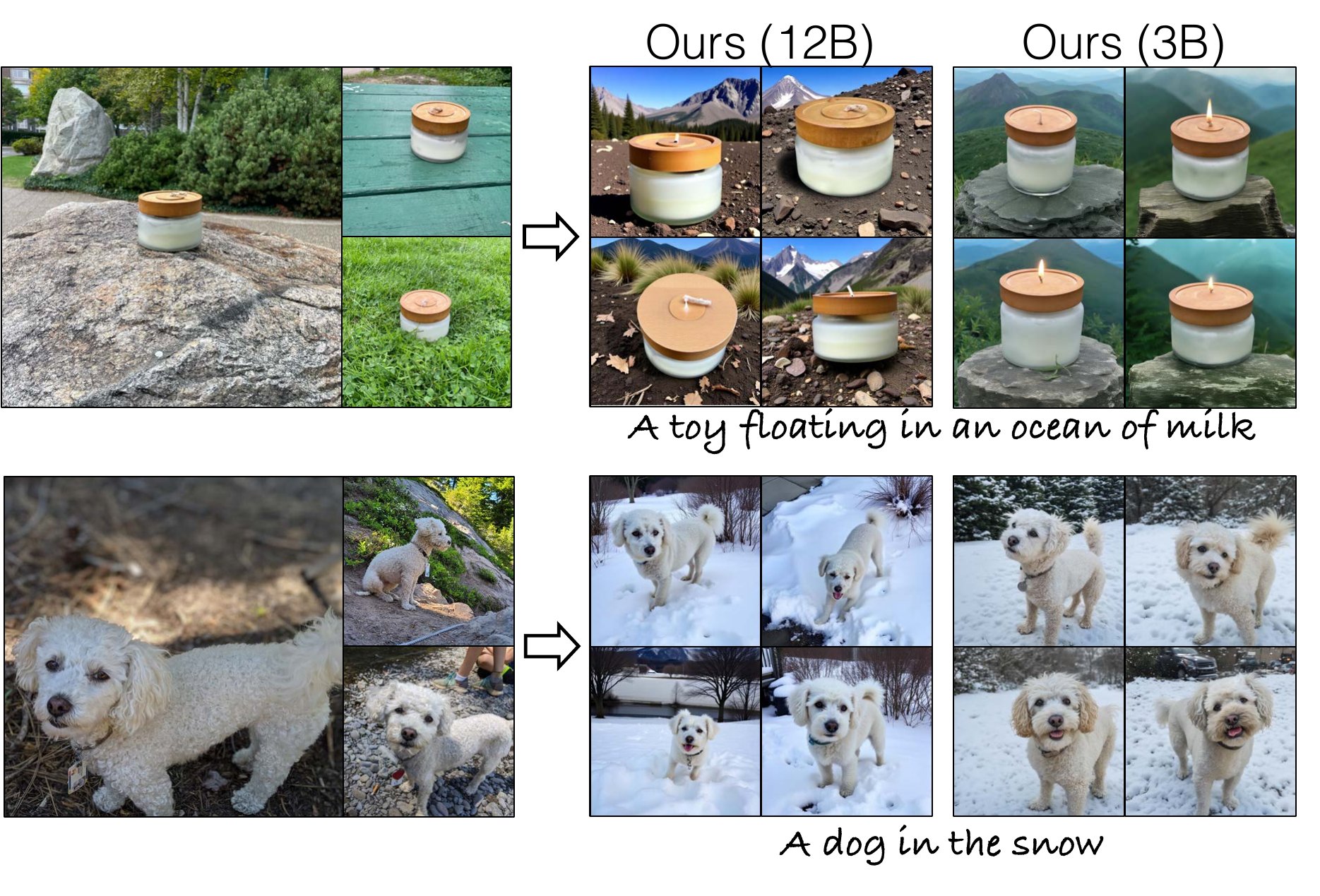}
    \vspace{-25pt}
    \caption{\textbf{Comparison between Ours (12B) and Ours (3B)}. Ours (12B) model, fine-tuned from FLUX, leads to higher diversity in object viewpoint and background, while trained on the same dataset, compared to Ours (3B) fine-tuned from a diffusion U-Net model.
    }
    \lblfig{sdxl_vs_flux} 
\end{figure}

\begin{table}[!t]
\centering
\setlength{\tabcolsep}{5pt}
\resizebox{\linewidth}{!}{
\begin{tabular}{l cc cc c}
\toprule
\multirow{3}{*} \textbf{Method}
& \multicolumn{2}{@{} c}{\textbf{MDINOv2-I}$\uparrow$} 
& \textbf{CLIPScore} $\uparrow$ 
& \textbf{TIFA} $\uparrow$ 
&  \textbf{Geometric} $\uparrow$ \\
&  Background &  Property & & & \textbf{Score} \\
& change prompt & change prompt & & \\
\midrule
\textbf{1-input} & & \\
IPAdapter Plus~\cite{ye2023ip}   & 0.618 & 0.626 &  0.261 & 0.569 & 0.595 \\
Emu-2~\cite{Emu2}  & 0.604 & 0.619 & 0.284 &  0.701 & 0.655 \\
Ours (3B)  &  0.645 & 0.609 & \textbf{0.315} & \textbf{0.809} & \textbf{0.712} \\
\hline 
\textbf{3-input} \\
Ours (3B) & \textbf{0.689}  & \textbf{0.666} & 0.304 & 0.749 & \textbf{0.712} \\
\bottomrule
\end{tabular}
}
\vspace{-8pt}
\caption{ \textbf{Results on CustomConcept101~\cite{kumari2023multi}}. Our method outperforms both Emu-2~\cite{Emu2} and IP-Adapter~\cite{ye2023ip} on the overall Geometric Score~\cite{yan2023motion} metric while being on par regarding image alignment. The Geometric Score is computed by taking the geometric mean of MDINOv2-I and TIFA scores, both of which are in the 0-1 range.  
}

\label{tbl:customconcept101}
\end{table}

\begin{figure*}[t]
    \centering
    \includegraphics[width=\linewidth]{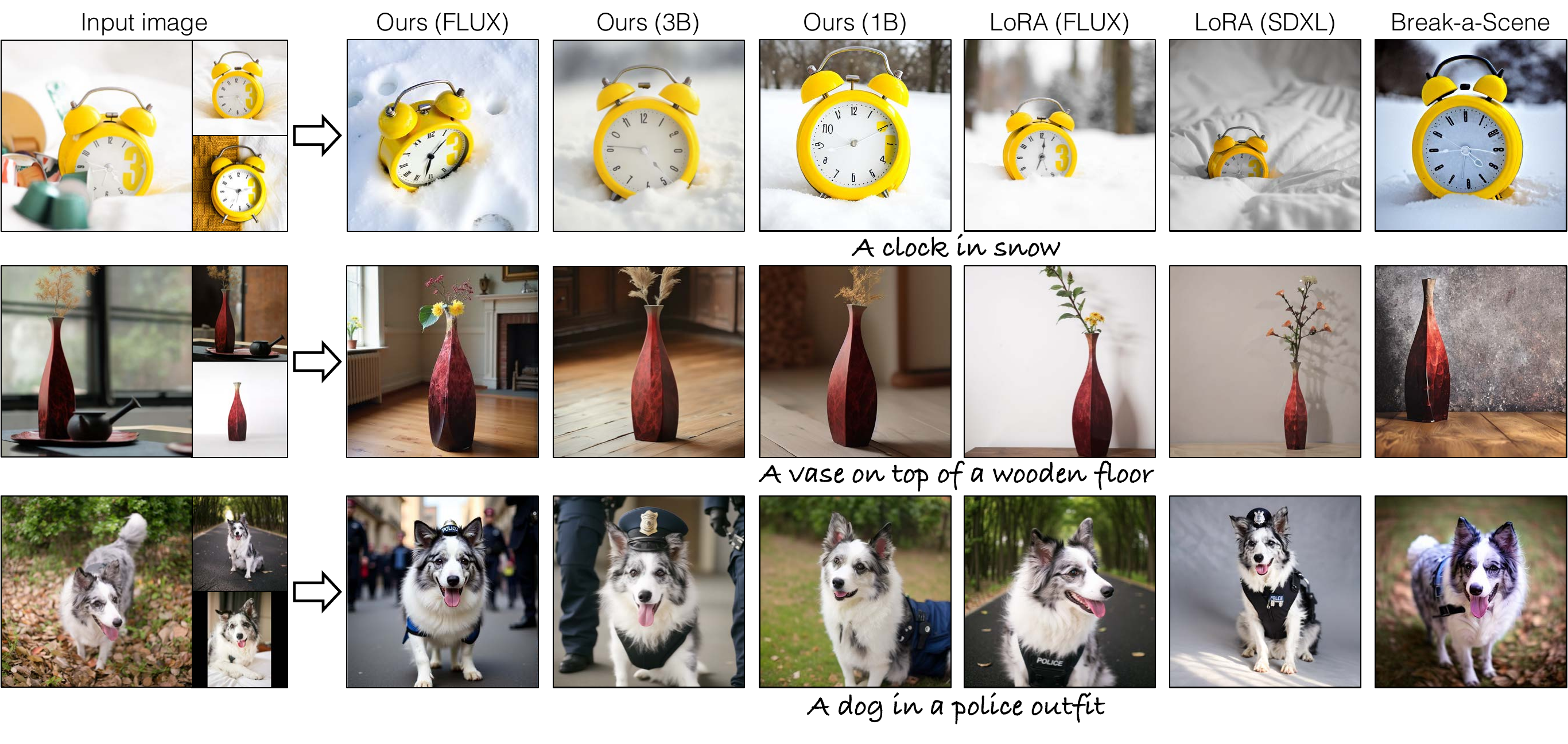}
    \vspace{-20pt}
    \caption{{\textbf{Comparison with optimization-based methods} We compare different models fine-tuned with our method and dataset, SynCD, to LoRA~\cite{loraimplementation} fine-tuned with same three reference image and Break-a-Scene~\cite{avrahami2023break} given 1-input image. Break-a-Scene can sometimes ignore the text prompt, e.g., {\menlo A dog in police outfit } in the last row. LoRA follows the text prompt but can still overfit to the reference image background, whereas our method follows the text prompt better while having on-par image alignment (e.g., $1^{\text{st}}$ column vs. $4^{\text{rth}}$ column). Break-a-Scene input is the first image from the $1^{\text{st}}$ column, and all other methods use the three images as input during training or inference.
    }}
    \lblfig{tuning_based_samples1} 
    \vspace{-15pt}
\end{figure*}

\begin{figure}[t]
    \centering
    \includegraphics[width=\linewidth]{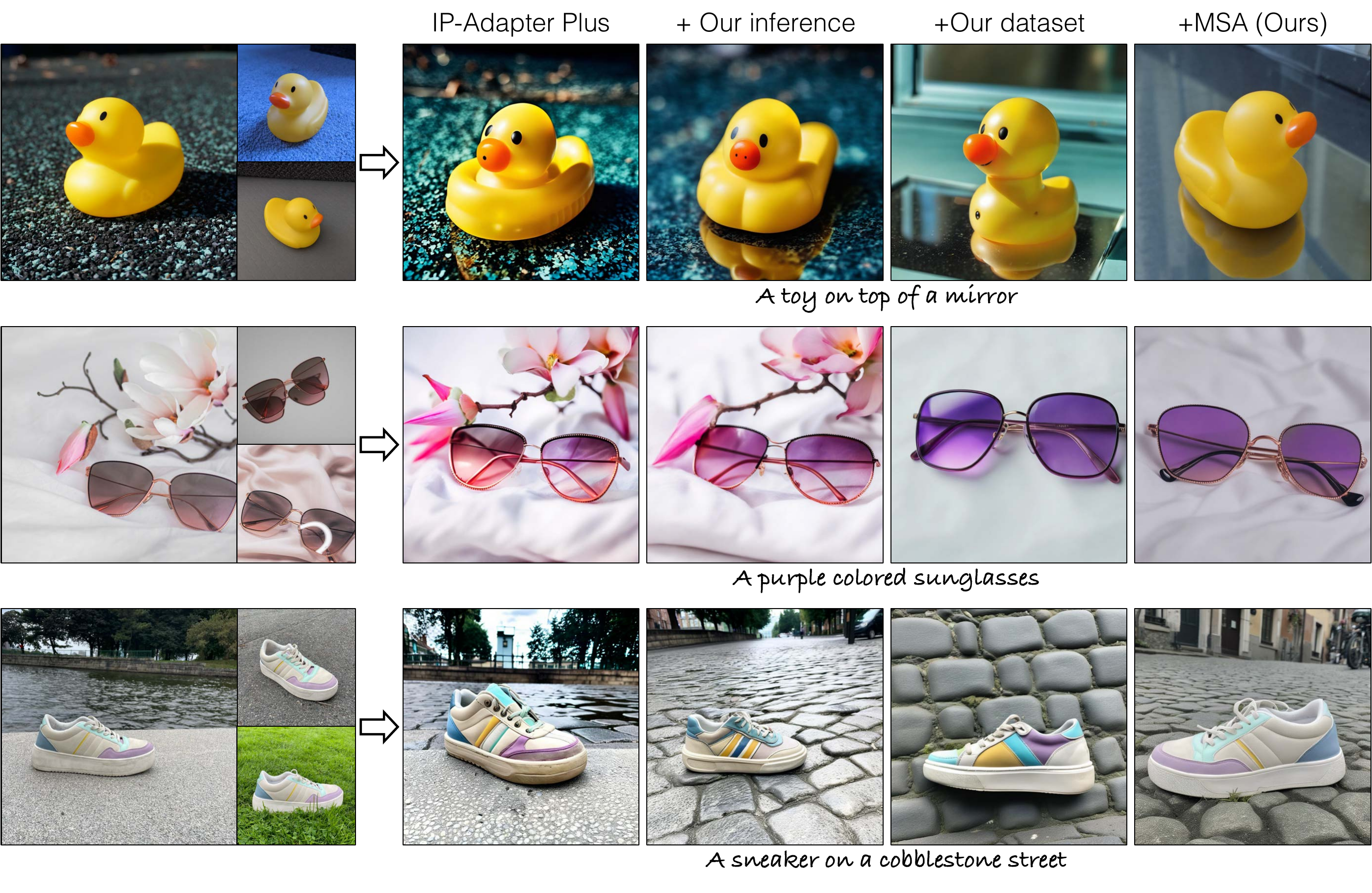}
    \vspace{-20pt}
    \caption{\textbf{Qualitative samples of ablation studies}. \textit{$1^{\text{st}}$ column}: Vanilla IP-Adapter Plus baseline samples. \textit{$2^{\text{nd}}$ column}: Modifying the inference technique to ours leads to higher text alignment with minor effect on image alignment, e.g., glass-like surface in $1^{\text{st}}$ row. \textit{ $3^{\text{rd}}$column}: Further fine-tuning on our dataset improves text prompt following, though with a decrease in object identity. \textit{ $4^{\text{rth}}$column}: Finally, having a Masked Shared Attention design for conditioning on multiple reference images improves the object identity without hurting text alignment. Please zoom in for details.
    }
    \vspace{-10pt}
    \lblfig{model_ablation_fig} 
\end{figure}

\begin{figure}[!t]
    \centering
    \includegraphics[width=\linewidth]{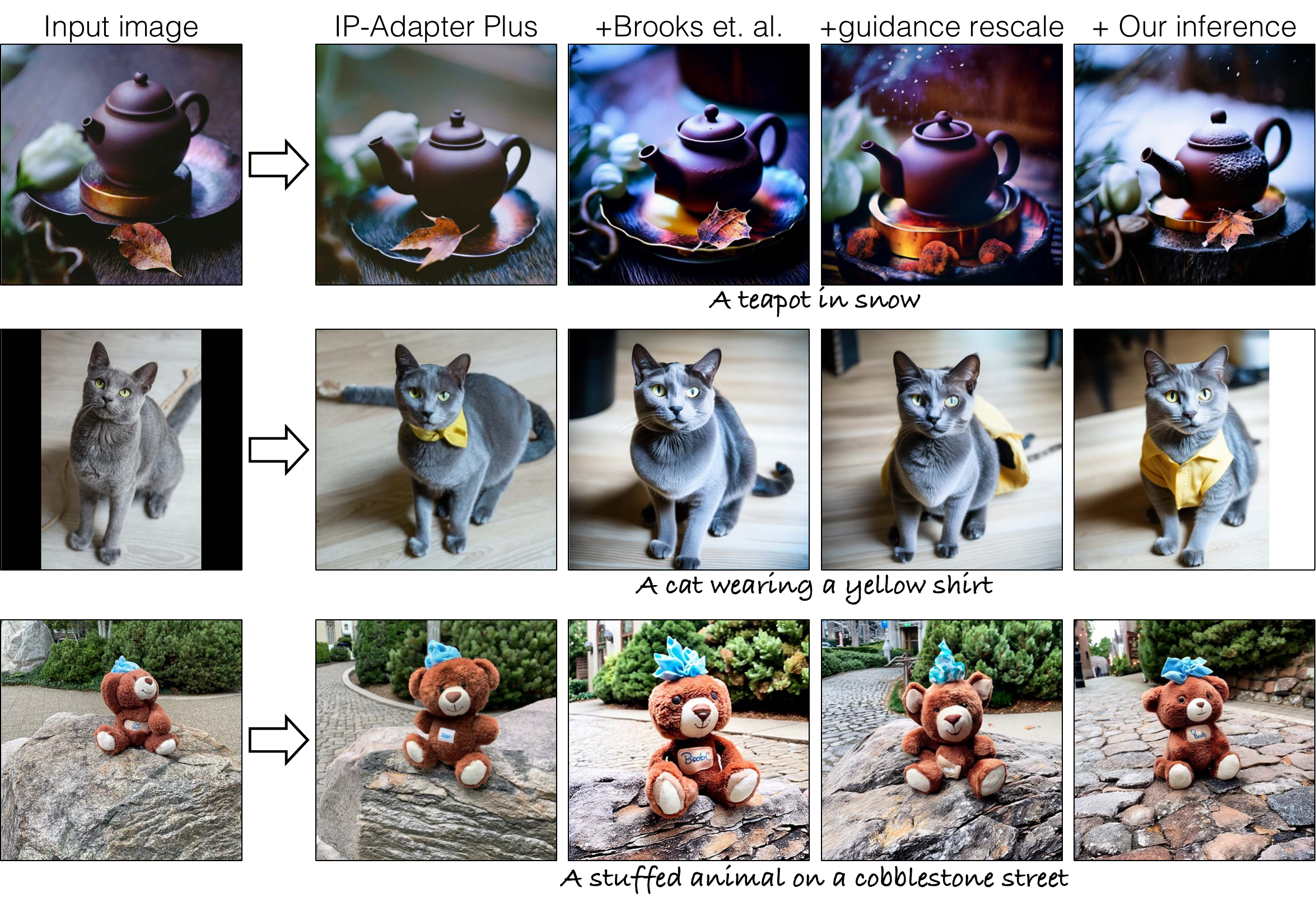}
    \vspace{-20pt}
    \caption{{\textbf{Qualitative comparison of our inference.} Our modified inference technique helps increase text alignment while minimally affecting the object identity. In comparison, the inference technique of Brooks~\etal~\cite{brooks2023instructpix2pix} or additional guidance rescale~\cite{lin2024common} has less effect on the final outputs. Please zoom in for details.
    }}
    \lblfig{inference_supp} 
\end{figure}

\begin{table}[!t]
\centering
\setlength{\tabcolsep}{5pt}
\resizebox{\linewidth}{!}{
\begin{tabular}{ll cc cc  c}
\toprule
& \multirow{3}{*} \textbf{Method}
& \multicolumn{2}{@{} c}{\textbf{MDINOv2-I}$\uparrow$} 
& \textbf{CLIP}$\uparrow$ & \textbf{TIFA} $\uparrow$ 
&  \textbf{Geometric} $\uparrow$ \\
&  & Background &  Property &\textbf{Score} & \textbf{Score} & \textbf{Score} \\
 &  & change prompt & change prompt & & \\

\midrule
\multirow{4}{*}{1-input} & Break-a-Scene~\cite{avrahami2023break}  & 0.765&0.752 & 0.304 & 0.823 & 0.791 \\
& Ours (1B) & 0.744 & 0.671  & 0.310 &  0.850 & 0.781 \\
& Ours (3B)  & 0.777 & 0.708 & 0.319 & 0.898 
& \textbf{0.825} \\
& Ours (12B)  & 0.749 & 0.732 & 0.309 &  0.751 &  0.734\\ 
\hline
\multirow{5}{*}{3-input} & LoRA (SDXL)~\cite{loraimplementation} &0.795&0.776 & 0.303  & 0.760 & 0.774 \\
& LoRA (FLUX) & \textbf{0.822} &  0.815 & 0.293 & 0.761 & 0.790 \\
&  Ours (1B) & 0.806 & 0.773 & 0.303 & 0.830  & 0.801 \\
& Ours (3B) & \textbf{0.822} & 0.789 & \textbf{0.313} &  \textbf{0.863}  & \textbf{0.838} \\
& Ours (12B)  & 0.763 & 0.744 & 0.305 &  0.781 & 0.765\\
\bottomrule
\end{tabular}
}
\vspace{-8pt}
\caption{ \textbf{Comparison with optimization-based methods}. Our method remains competitive against optimization-based methods, with better text alignment and comparable image alignment, as also shown qualitatively in \reffig{tuning_based_samples1}. }

\label{tbl:tuningbased_metrics}
\vspace{-5pt}
\end{table}

\section{Ablation Study}\lblsec{appendix_ablation}

\myparagraph{Model ablation.} 
Here, we show more qualitative and quantitative comparisons of the ablation experiments reported in Section 5.2 of the main paper, i.e., when gradually adding our inference, dataset, and shared attention mechanism to the IP-Adapter baseline. \reffig{model_ablation_fig} shows a qualitative comparison when fine-tuning the IP-Adapter on our dataset and subsequently adding Masked shared attention (MSA). MSA helps the model capture fine details, e.g., the specific color pattern of the shoe in the last row. For our modified inference, \reftbl{inference_supp} compares it with the default inference of Brooks~\etal ~\cite{brooks2023instructpix2pix} and guidance rescale~\cite{lin2024common} on the IP-Adapter Plus~\cite{ye2023ip} baseline, with the same text and image guidance scale. The default inference technique of Brooks~\etal ~\cite{brooks2023instructpix2pix} and adding guidance rescale to it do not affect the final performance significantly. With our normalization technique, the text alignment improves with a comparatively minor drop in image alignment. The sample comparisons in \reffig{inference_supp} also show the same trend.

\begin{table}[!t]
\centering
\setlength{\tabcolsep}{5pt}
\resizebox{\linewidth}{!}{
\begin{tabular}{l cc cc c}
\toprule
\multirow{3}{*} \textbf{Method}
& \multicolumn{2}{@{} c}{\textbf{MDINOv2-I}$\uparrow$} 
& \textbf{CLIPScore} $\uparrow$ 
& \textbf{TIFA} $\uparrow$ 
&  \textbf{Geometric} $\uparrow$ \\
&  Background &  Property & & & \textbf{Score} \\
& change prompt & change prompt & & \\
\midrule
IPAdapter Plus   & \textbf{0.744} & \textbf{0.737} & 0.270 & 0.615 & 0.675 \\
+ Guidance rescale~\cite{lin2024common} (0.6)  &0.722 & 0.699 & 0.276 & 0.707 & 0.710 \\
+ Vanilla Img + Text~\cite{brooks2023instructpix2pix} & 0.722 &  0.711 & 0.270 & 0.681 & 0.699 \\
+ Our inference & 0.719 & 0.668 & \textbf{0.298} &  \textbf{0.816} & \textbf{0.756} \\
\bottomrule
\end{tabular}
}
\vspace{-8pt}
\caption{ \textbf{Our inference}. We compare our inference technique with vanilla image and text guidance technique~\cite{brooks2023instructpix2pix} as well as guidance rescale~\cite{lin2024common} with the same inference hyperparameters across all. 
}

\label{tbl:inference_supp}
\end{table}

\begin{table}[!t]
\centering
\setlength{\tabcolsep}{5pt}
\resizebox{\linewidth}{!}{
\begin{tabular}{ll cc cc c}
\toprule
& \multirow{3}{*} \textbf{Method}
& \multicolumn{2}{@{} c}{\textbf{MDINOv2-I}$\uparrow$} 
& \textbf{CLIP}$\uparrow$ & \textbf{TIFA} $\uparrow$ 
&  \textbf{Geometric} $\uparrow$ \\
&  & Background &  Property &\textbf{Score} & \textbf{Score} & \textbf{Score} \\
 &  & change prompt & change prompt & & \\
\midrule
\multirow{2}{*}{1-input} &  Ours (12B) & \textbf{0.749} & \textbf{0.732} & 0.309 & \textbf{0.853} &  \textbf{0.795}   \\
& w/ OminiControl~\cite{tan2024ominicontrol}  & 0.739 & 0.717 & \textbf{0.313} & \textbf{0.853} & 0.789 \\
\hline
\multirow{2}{*}{3-input} & Ours (12B) & \textbf{0.778} & \textbf{0.771} & \textbf{0.306} & \textbf{0.786} & \textbf{0.780} \\
& w/ OminiControl~\cite{tan2024ominicontrol}  & 0.764 & 0.760 & 0.303 & 0.761 & 0.760 \\
\bottomrule
\end{tabular}
}
\vspace{-8pt}
\caption{ \textbf{SynCD vs. OminiControl dataset} using the same FLUX model fine-tuning and inference protocols. Model trained on our dataset performs better than OminiControl dataset in terms of image alignment, specifically it can better accommodate multiple reference images, as also shown in \reffig{ominicontroldata_fig}, thus highlighting the advantage of our dataset generation pipeline.
}

\label{tbl:omini_data}
\end{table}

\begin{figure}[t]
    \centering
    \includegraphics[width=\linewidth]{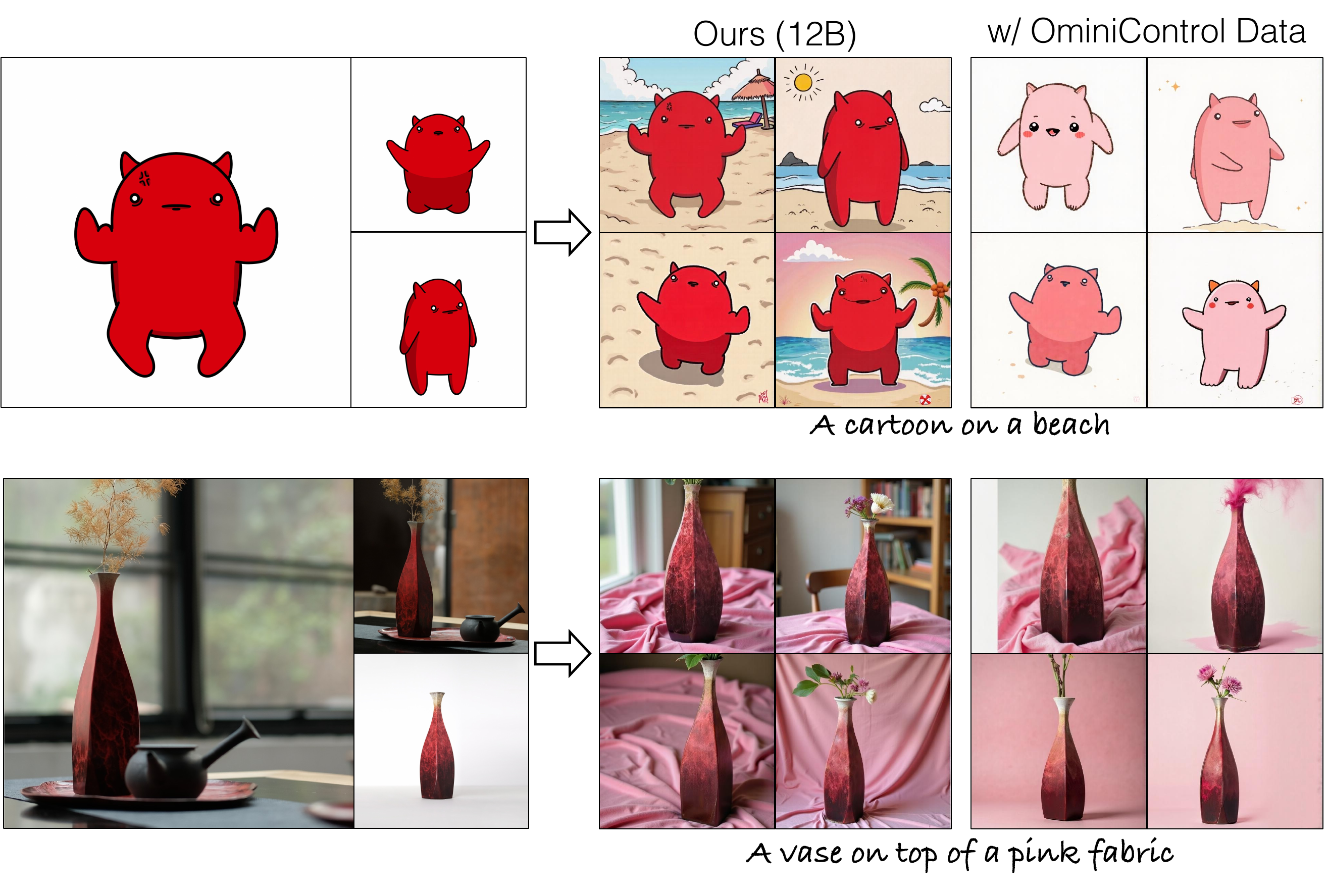}
    \vspace{-25pt}
    \caption{\textbf{Comparison with OminiControl data}. Fine-tuning FLUX on our dataset, consisting of $2-3$ images per object, leads to better generalization on using multiple reference images during inference. Comparatively, the FLUX model fine-tuned on the OminiControl~\cite{tan2024ominicontrol} dataset struggles with multiple reference images as input.
    }
    \vspace{-10pt}
    \lblfig{ominicontroldata_fig} 
\end{figure}

\begin{figure}[!t]
    \centering
    \includegraphics[width=0.95\linewidth]{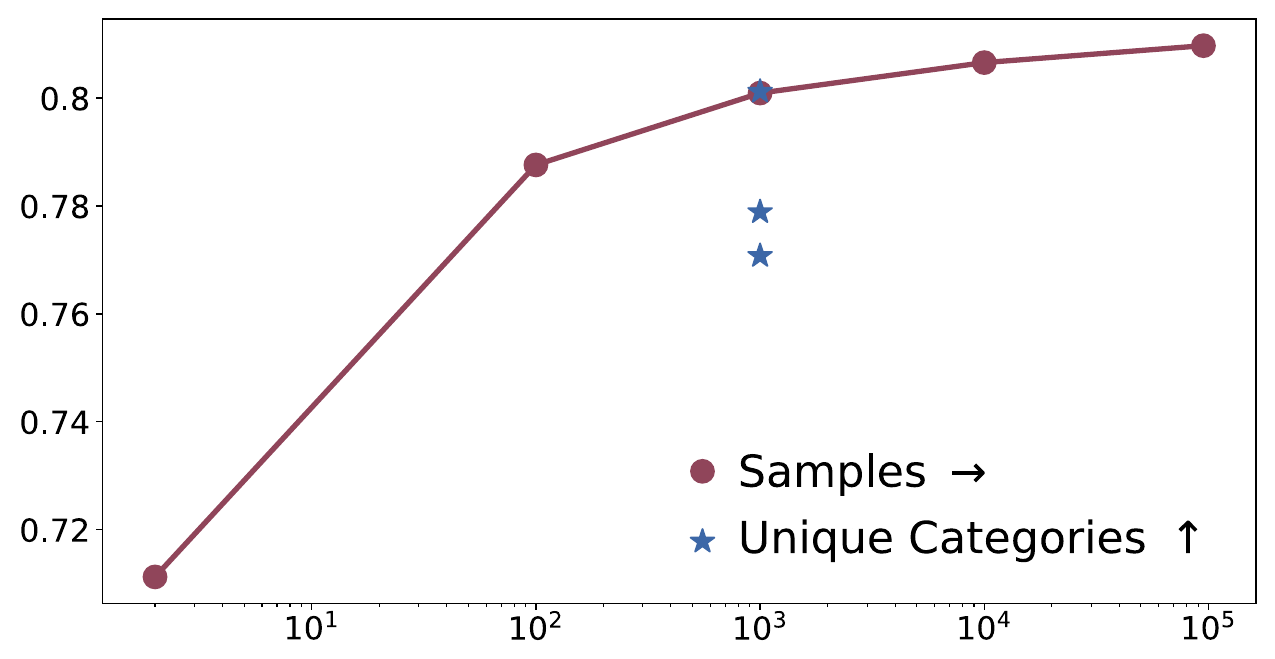}
    \vspace{-10pt}
    \caption{{\textbf{Dataset ablation.} We plot the MDINOv2-I metric with increased sample size and category diversity. Given the same sample size of 1K objects, increasing category diversity from $3$ to $16$ and $200$ gradually improves image alignment. 
    }}
    \lblfig{data_cat_plot} 
\end{figure}

\begin{figure*}[!t]
    \centering
    \includegraphics[width=0.88\linewidth]{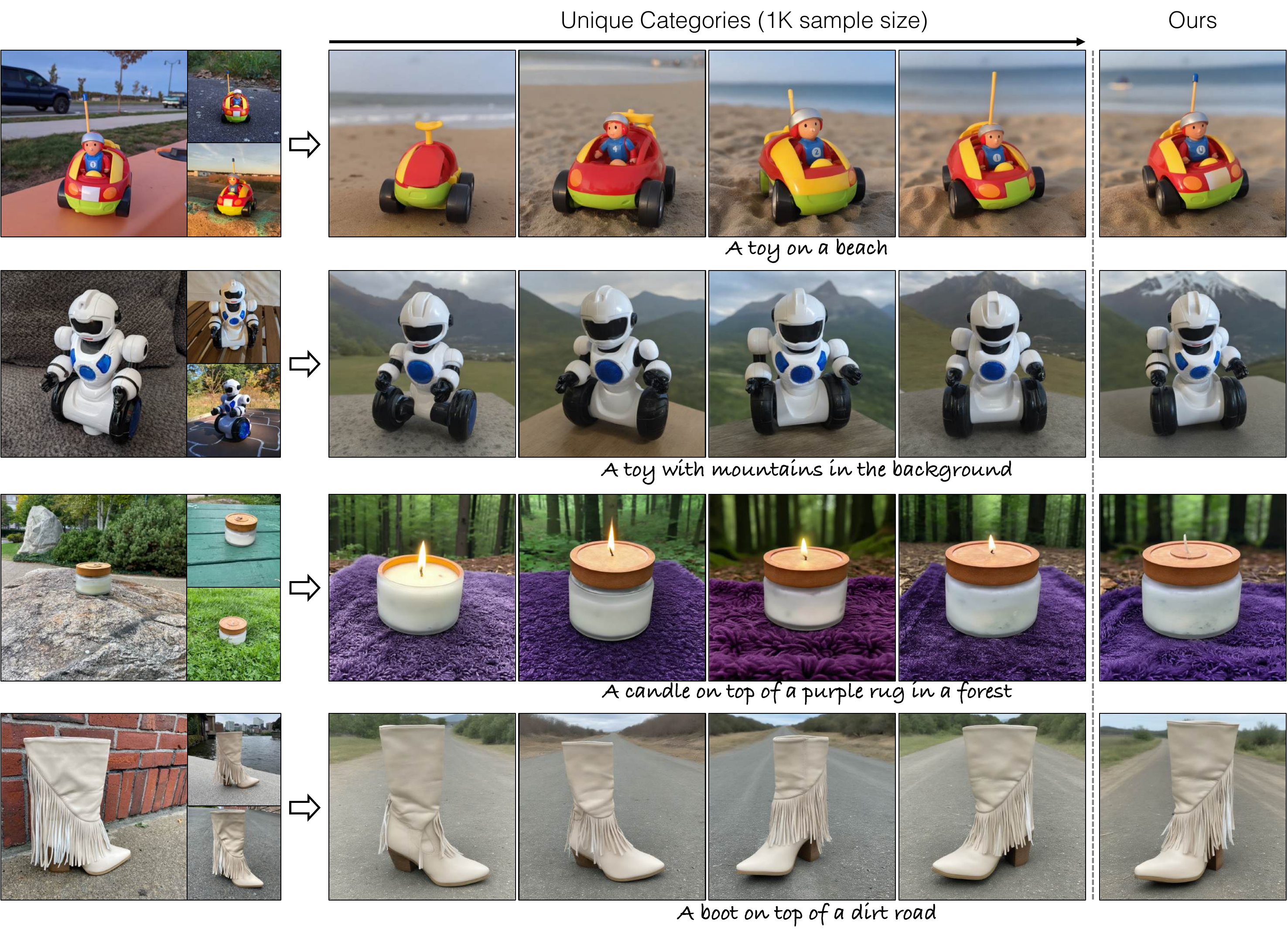}
    \vspace{-15pt}
    \caption{{\textbf{Effect of dataset category diversity on performance.} As we increase the number of unique categories from  $1$ to $3$ to $16$ and $200$ (with a fixed sample size of 1K), performance improves with the model capturing finer details of the object, e.g., the unique pattern in front of the toy car in $1^{\text{st}}$ row or the frills of the boot in $4^{\text{rth}}$ row. \textbf{Please zoom in for details.}
    }}
    \lblfig{dataset_ablation}
    \vspace{-10pt}
\end{figure*}

\begin{figure*}[!t]
    \centering
    \includegraphics[width=0.88\linewidth]{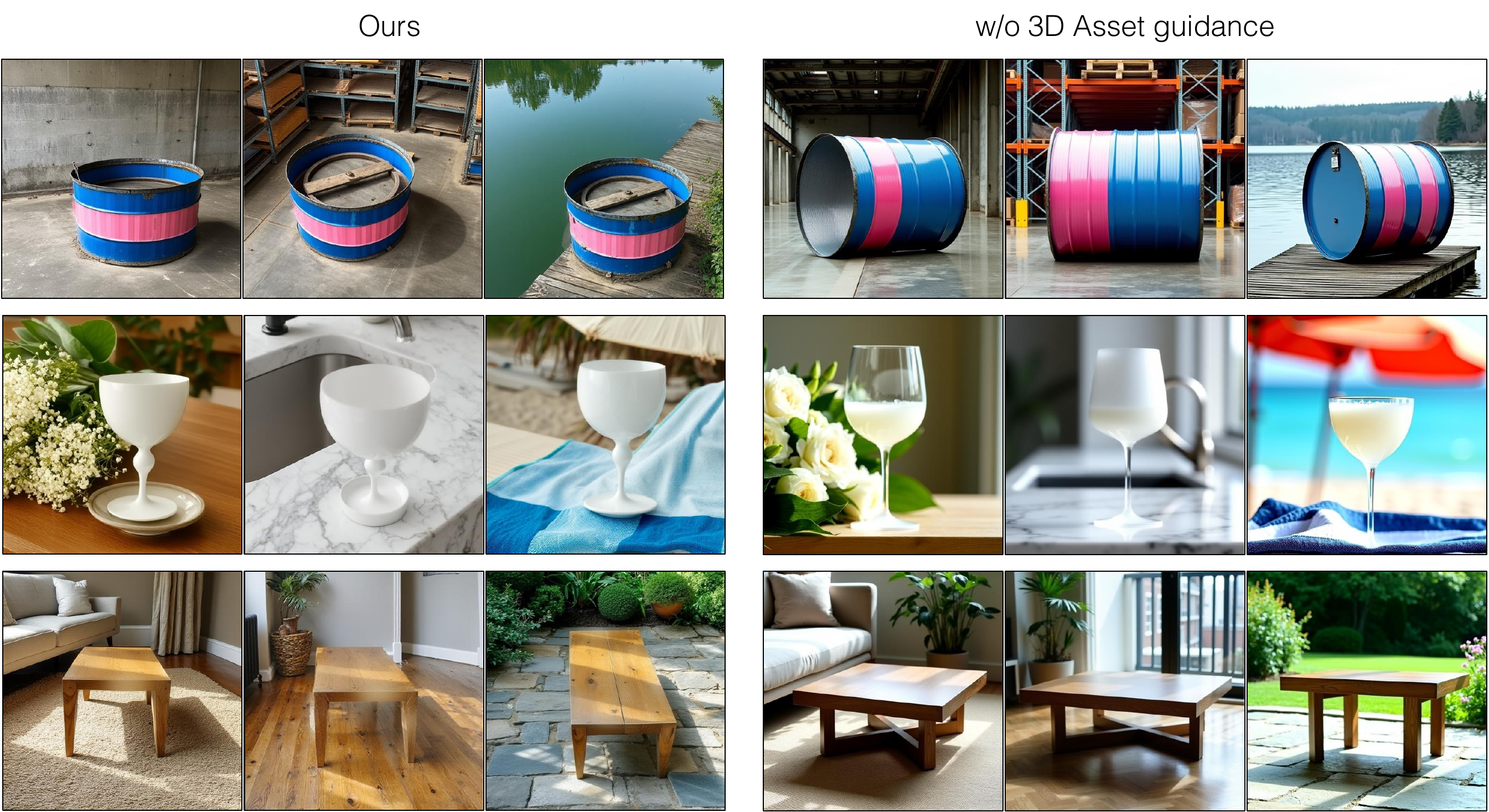}
    \vspace{-8pt}
    \caption{\textbf{Rigid object generation w/ vs. w/o 3D Asset guidance.} We compare our final rigid object generation results with that of removing 3D asset guidance and only using MSA. Removing depth and warping guidance from the dataset generation pipeline reduces multi-view and shape consistency. 
    }
    \lblfig{dataset_ablation2}
\end{figure*}

\myparagraph{Comparison to OminiControl~\cite{tan2024ominicontrol} dataset.} 
OminiControl is another concurrent work that introduces a synthetic dataset generated from existing text-to-image models like FLUX~\cite{flux} for customization. They rely on prompting alone to create two side-by-side images of the same object. In contrast, our method also relies on more explicit constraints like depth and cross-view correspondence from Objaverse~\cite{deitke2023objaverse}. Here, we compare our dataaset to OminiControl while keeping training and inference protocols the same as ours. \reftbl{omini_data} Shows this comparison across different metrics and \reffig{ominicontroldata_fig} shows qualitative samples. Our method better incorporates the object identity while still following the text prompt, e.g., the pink fabric in $2^{\text{nd}}$ column of \reffig{ominicontroldata_fig}.  Comparatively, the model trained on the OminiControl dataset can overfit to the reference images. Since our dataset consists of $2-3$ images per object, we use more than $1$ reference image during training, which leads to better generalization on using multiple reference images during inference as well.

\myparagraph{Dataset diversity.} 
We examine the impact of category diversity on the final model performance by creating various subsets of the data with more diverse categories using Objaverse tags for each asset. As shown in \reffig{data_cat_plot}, we plot the image alignment MDINOv2-I metric for different subsets while keeping the sample size fixed to 1K, and higher category diversity leads to better performance. Similarly, given the same category diversity, performance quickly plateaus, thus suggesting that ensuring high diversity is a critical factor in the final model performance.

\myparagraph{Rigid object generation.} \reffig{dataset_ablation2} here shows that only having MSA for rigid objects fails to maintain the same shape and multi-view consistency across different views. Whereas guiding the generation using 3D assets from datasets like Objaverse~\cite{deitke2023objaverse} leads to more consistent objects.

\section{Implementation details}
\lblsec{details}

\subsection{Dataset Generation Details}\lblsec{details_dataset_gen}

\myparagraph{LLM instruction details.}
To get a set of prompts in our dataset generation, we use Instruction-tuned LLama3~\cite{dubey2024llama}. The input instruction to the LLM always consists of the prompt shown below, which is modified from Esser~\etal~\cite{esser2024scaling}: 
\begin{tcolorbox}
\textbf{Role}: system, \textbf{Content}: You are a language model expert in suggesting image captions for different object categories.\\
\textbf{Role}: user, \textbf{Content}: suggest ten captions for images of a [\textcolor{blue}{object description/ category}]. The caption should provide a [\textcolor{blue}{TASK}]. DO NOT add any unnecessary adjectives or emotional words in the caption. Please keep the caption factual and terse but complete. DO NOT add any unnecessary speculation about the things that are not part of the image, such as ``the image is inspiring to viewers'' or ``seeing this makes you feel joy''. DO NOT add things such as ``creates a unique and entertaining visual'', as these descriptions are interpretations and not a part of the image itself. The description should be purely factual, with no subjective speculation.
\end{tcolorbox}

Where in the case of rigid object generation, we provide the \textcolor{blue}{object description} from CAP3D~\cite{luo2024scalable} and the \textcolor{blue}{TASK} is ``a description of the background''. We also provide two sample descriptions, as shown below: 
\begin{tcolorbox}
Follow this guidance for the captions: 
\begin{enumerate}
    \item Generate captions of [\textcolor{blue}{object description}] in different backgrounds and scenes. 
    \item Generate captions of [\textcolor{blue}{object description}] with another object in the scene. 
\end{enumerate}

Example captions for ``White plastic bottle'' are:
\begin{enumerate}
    \item A white plastic bottle on a roadside cobblestone with stone bricks.
    \item A white plastic bottle is placed next to a steaming cup of coffee on a polished wooden table.
\end{enumerate}

Example captions for a ``blue truck'' are:
\begin{enumerate}
    \item A blue tank in a military storage facility with metal walls.
    \item A blue tank on a desert battlefield ground, with palm trees in the background.
\end{enumerate}

\end{tcolorbox}

In the case of deformable object generation, we prompt the LLM once, with the \textcolor{blue}{category} name, e.g., cat, and \textcolor{blue}{TASK} as ``detailed visual information of the category, including color and subspecies''. We append the below instruction as well to the LLM: 

\begin{tcolorbox}
Example caption descriptions for the category ``cat'':
\begin{enumerate}
    \item  The Siamese cat has blue almond-shaped eyes and cream-colored fur with dark chocolate points on the ears, face, paws, and tail. 
    \item The white fluffy Maine Coon cat with a long and bushy tail spread out beside it, and its thick fur has a mix of brown, black, and white stripes. 
    \item The Bengal cat with a marbled coat features a pattern of vivid orange and black spots.
\end{enumerate}
\end{tcolorbox}

We prompt the LLM again with the same \textcolor{blue}{category} name and \textcolor{blue}{TASK} as `` a description of the background''. We append the below instruction as well to the LLM: 
 \begin{tcolorbox}
Follow this guidance for the captions: 
\begin{enumerate}
    \item Generate captions of [\textcolor{blue}{category}] in different backgrounds and scenes. 
    \item Generate captions of [\textcolor{blue}{category}] with another object in the scene. 
    \item Generate captions of [\textcolor{blue}{category}] with different stylistic representations. 
\end{enumerate}

Example captions for the category ``cat'' are:
\begin{enumerate}
    \item Photo of a cat playing in a garden. The garden is filled with wildflowers. 
    \item A cat is sitting beside a book in a library. 
    \item Painting of a cat in watercolor style.
\end{enumerate}
\end{tcolorbox}

\myparagraph{Masked Shared Attention (MSA).} When performing MSA in DiT-based text-to-image models, we modify the rotational positional encoding~\cite{su2024roformer} to be $N H \times W$ image resolution for generating the $N$ images of $H \times W$ resolution. Further, during sampling, each image attends to everything in the other image at the first time step, and the mask is then used in subsequent time steps. 
More specific details related to rigid and deformable object generation are provided below. 

\myparagraph{Rigid object generation.} We select approximately $75K$ assets from the Objaverse dataset~\cite{deitke2023objaverse}, a subset of LVIS~\cite{gupta2019lvis} and high-quality assets shared by Tang~\etal~\cite{tang2025lgm}. To describe each asset, we utilized detailed prompts provided by Cap3D~\cite{luo2024scalable}. We render the asset from a uniformly sampled camera viewpoint in the upper hemisphere with a maximum elevation of $70$ degrees, and for each set, select three views with a minimum $10\%$ pairwise overlap in the rendered images. We then pre-calculate the cross-view pixel correspondence between them, which is used later for feature warping in the dataset generation pipeline. 

For generating samples in each set, we use ground truth rendered depth images as input to the depth-conditioned FLUX model~\cite{fluxdepth} along with negative prompts, such as {\menlo 3d render, low resolution, blurry, cartoon}. We apply feature warping to the first $20\%$ of denoising timesteps. Sampling is performed with $30$ steps of Euler Scheduler~\cite{esser2024scaling} at $512$ resolution, using a depth guidance of $10.0$ and a classifier-free guidance scale of $2.5$.  

\myparagraph{Deformable object generation.} 
In the case of deformable objects, we compute the mask of the foreground object region via text cross-attention~\cite{hertz2022prompt}, which is updated at every diffusion timestep. This is then used in the Masked Shared Attention (MSA) to enable foreground object regions to attend to each other in each set. Additionally, once the images are generated, we remove the detailed object descriptions from the prompt in the final dataset. The images are generated with $50$ sampling timesteps and standard text guidance of $3.5$ at $1$K resolution.

\begin{figure*}[!t]
    \centering
    \includegraphics[width=\linewidth]{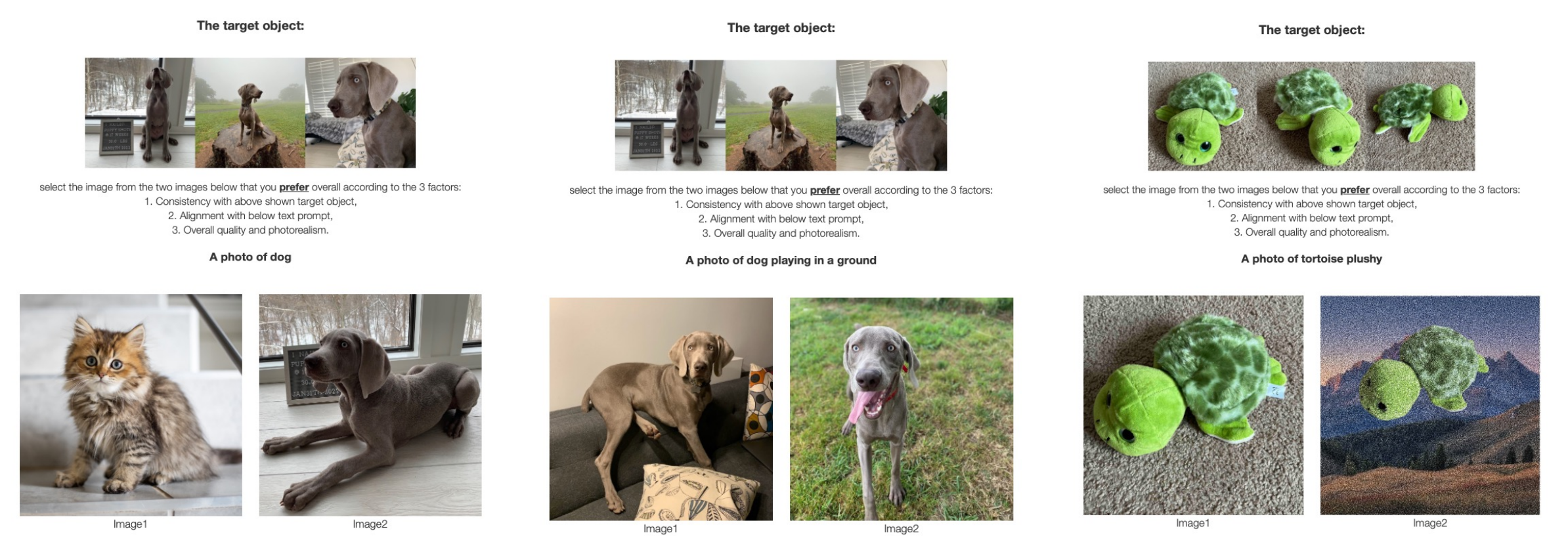}
    \vspace{-20pt}
    \caption{{\textbf{Sample practice test for human preference study.} We show $3$ practice questions to each participant that test their ability to select the images based on the three criteria that we care about, i.e., identity preservation or image alignment, text alignment, and overall quality. 
    }}
    \lblfig{practice_fig} 
    \vspace{-10pt}
\end{figure*}

\begin{figure}[!t]
    \centering
    \includegraphics[width=\linewidth]{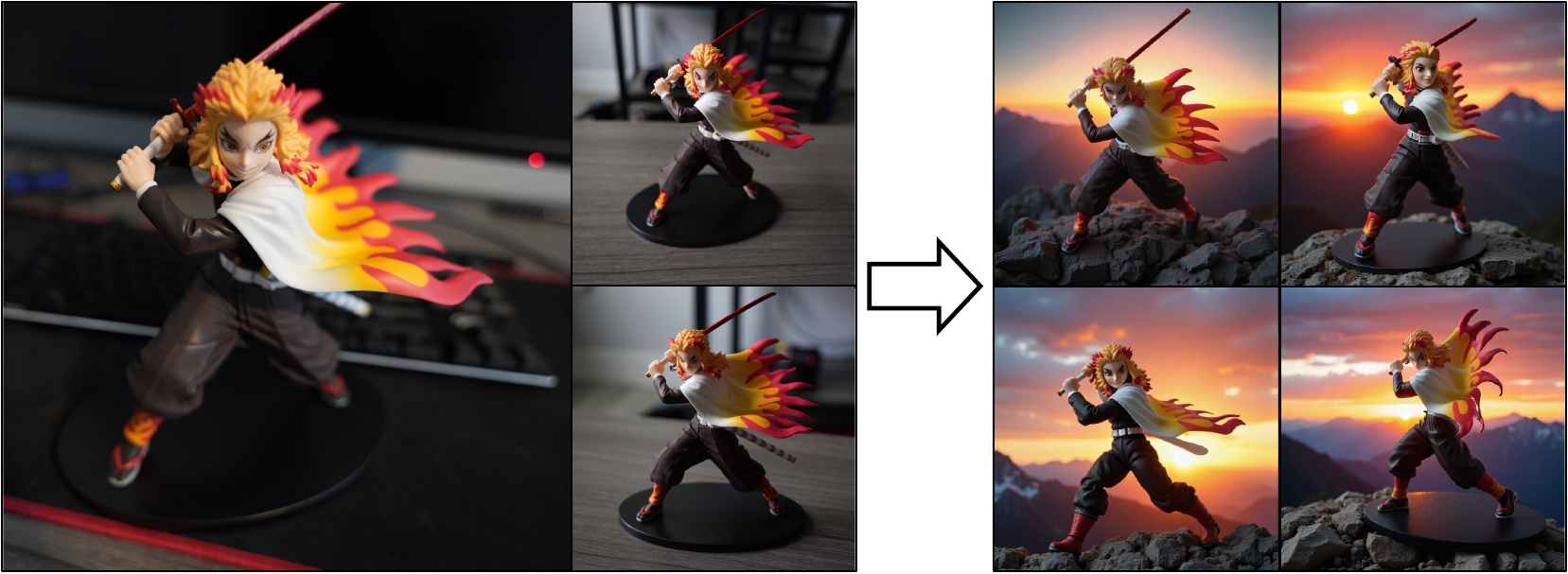}
    \vspace{-10pt}
    \caption{{\textbf{Limitation.} Our method can have limited variations in viewpoint and pose if the input reference images are also all in a similar pose. Images generated with Ours (12B) model.
    }}
    \lblfig{limitation_fig} 
    \vspace{-10pt}
\end{figure}

\begin{figure}[!t]
    \centering
    \includegraphics[width=0.85\linewidth]{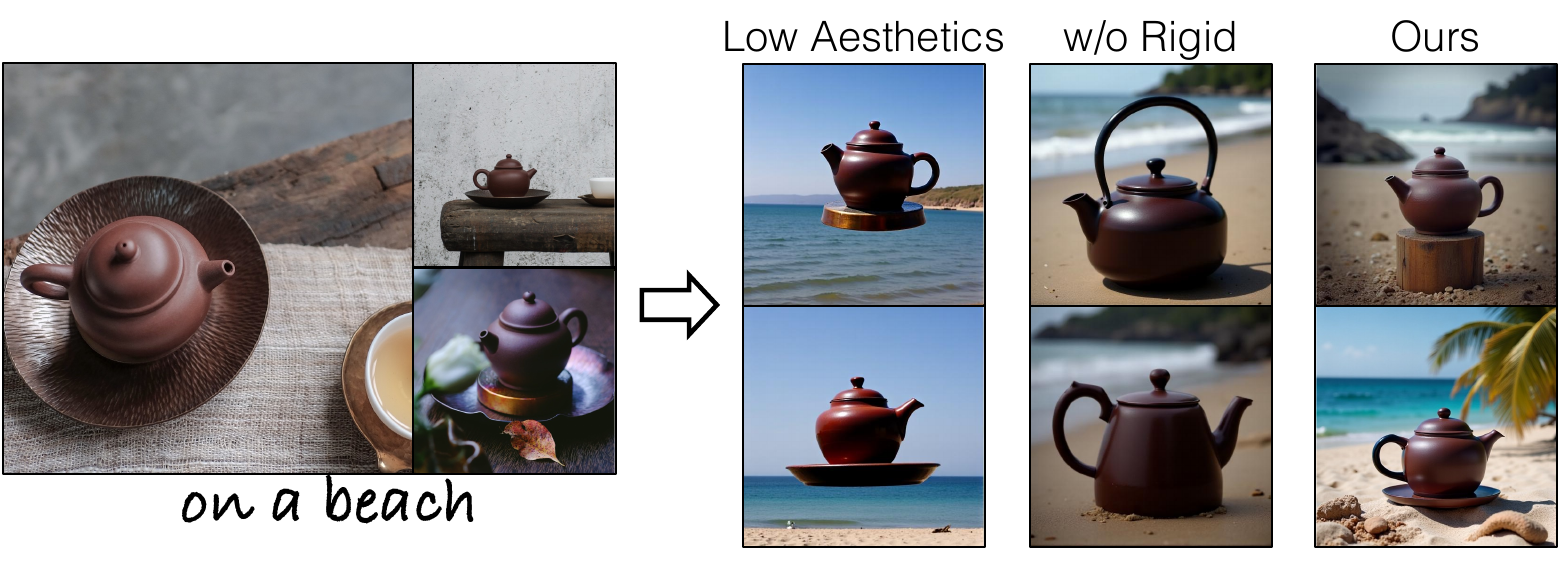}
    \caption{{\textbf{Impact of low-quality synthetic samples.} Model performance with low-quality subset is serverly affects compared to our final filtered dataset. This highlights the importance of dataset filtering when using synthetic samples for model fine-tuning.
    }}
    \lblfig{badsyncd} 
\end{figure}

\subsection{Our Method Details}\lblsec{our_method_details}
\myparagraph{Training.}
For Ours (3B) and Ours (1B), diffusion U-Net-based models, we train with a LoRA rank of $128$, batch size $32$, and learning rate $5\times 10^{-6}$ for $20K$ iterations. When extracting reference features in shared attention, we add the same timestep noise to reference images as the target. Our model is initialized from IP-Adapter Plus, which also conditions generation on CLIP features of an image via decoupled image cross-attention. 

For Ours (12B), fine-tuned from FLUX~\cite{flux}, we train with a LoRA rank of $32$, batch size $8$, and Prodigy optimizer~\cite{mishchenko2023prodigy} for $15K$ iterations. We do not use IP-Adapter initialization in this case; instead, we extract features of clean non-noisy reference images in the shared attention blocks. The RoPE~\cite{su2024roformer} is modified to be $(K+1)H \times W$, given $K$ reference images and the target noisy image. 

Finally, for both, training is done with a variable number of reference images, either $1$ or $2$, depending on the number of images in each set after filtering out low-quality samples.

\myparagraph{Inference.}
For Ours (3B) and Ours (1B), diffusion U-Net-based models, we sample using $50$ steps of the Euler Discrete scheduler~\cite{karras2022elucidating}. The text-guidance scale is set to $7.5$, and an adaptive image-guidance scale is used ($\lambda_{I}$ in Eq. 4 of the main paper), starting from $8.0$ for background change prompts and $6.0$ for property/shape change prompts and linearly increasing it by $5.0$ during the $50$ sampling steps. The IP-Adapter scale is always set to its default value of $0.6$. 

The inference time for sampling one image is $16$ and $29$ seconds, given $1$ and $3$ images as reference input, respectively, compared to $3$ seconds for the base pretrained model in $bfloat16$ on H100 GPU. The overhead is because of the longer sequence length in the masked shared attention with a dynamic mask, combined with making the forward call to the model twice at every step to extract reference features. 

For Ours (12B), we set the FLUX distilled guidance scale to its default value of $3.5$ and used classifier-free text and image guidance scale, $\lambda_{I}$ and $\lambda_{c}$ in  Eq. 4 of the main paper, as $1.0$ and $1.5$. The sampling is done with $30$ steps with default Flow matching Euler Discrete Scheduler~\cite{esser2024scaling}.

The inference time for sampling an image is $15$ and $25$ seconds, given $1$ and $3$ images as reference, respectively, compared to $3$ seconds for the base pretrained model in $bfloat16$ on H100 GPU. During shared attention, the target features attend to all foreground and background reference features. We do not observe a significant benefit of using masks in shared attention for the FLUX-based model, given the high computational overhead of using dynamic masks.

\subsection{Baselines}
Here, we mention the implementation details of baseline methods. For baselines with recommended hyperparameters, we always followed those while keeping the sampling step consistent across all to $30$ for FLUX~\cite{flux} -based models and $50$ for diffusion-based models. Similarly, the text guidance scale is $3.5$ and $7.5$ for the FLUX and diffusion-based models, respectively, unless otherwise mentioned. 

\myparagraph{OminiControl~\cite{tan2024ominicontrol}}. We used their open-source model based on FLUX-schnell~\cite{fluxschnell}. Following their paper, we replace category names in each text prompt with {\menlo ``this item''}. For sampling, we followed their recommended number of inference steps as $8$. 

\myparagraph{Kosmos-G~\cite{pan2023kosmos}.} We follow their open-source code to sample images on the DreamBooth evaluation dataset.

\myparagraph{BLIP Diffusion~\cite{li2023blip}.} According to the recommended technique, we modify each prompt to be an (image, category name, instruction) tuple where instruction is modified from the input prompt, e.g., ``{\menlo toy in a junle} ''$\rightarrow$ ``{\menlo in a jungle}'' or ``{\menlo a red toy}'' $\rightarrow$ ``{\menlo make it red}''. Additionally, we use the negative prompts provided in their open-source code. 

\myparagraph{IP-Adapter~\cite{ye2023ip}.} In the case of IP-Adapter~\cite{ye2023ip}, we use the IP-Adapter Plus with a U-Net-based diffusion model of the same parameter scale as Ours (3B). We use the recommended $0.6$ IP-Adapter scale.

\myparagraph{MoMA~\cite{song2024moma}.} We use their open-source code with the maximum strength parameter of $1$ for increased object identity preservation. 

\myparagraph{JeDi~\cite{zeng2024jedi}.} We use the generated images on the DreamBooth evaluation dataset shared by the authors. 

\myparagraph{Emu-2~\cite{Emu2}}. We use their open-source code with the recommended guidance of $3$. Additionally, as mentioned in their paper, we modify each prompt to be an (image, instruction) tuple where instruction is modified from the input prompt, e.g., ``{\menlo toy in a junle}'' $\rightarrow$ ``{\menlo in a jungle}'' or ``{\menlo a red toy}'' $\rightarrow$ ``{\menlo make it red}''.

\myparagraph{Break-a-Scene~\cite{avrahami2023break}.} We use the open-source code of Break-a-Scene and learn $2$ assets, one corresponding to the object and another for the background. During inference, we only use the learned asset for the object.

\myparagraph{LoRA~\cite{hu2021lora,loraimplementation}.} We follow the hyperparameters from the HuggingFace implementation~\cite{loraimplementation} and fine-tune a U-Net-based diffusion model of the same parameter scale as Ours (3B). Additionally, we enable class regularization with generated images to prevent overfitting, as suggested in DreamBooth~\cite{ruiz2022dreambooth}. 

\myparagraph{Guidance rescale.} We incorporate guidance rescale~\cite{lin2024common} in the image and text-guidance formulation of Brooks~\etal~\cite{brooks2023instructpix2pix} by first computing the final prediction $\epsilon_{\theta}$ using text- and image-guidance, then rescaling  $\epsilon_{\theta}$ with the standard deviation ratio of $\epsilon_{\theta}(\x^t, \{\x_i\}_{i=1}^K, \mathbf{c})$ and  $\epsilon_{\theta}$, similar to the guidance rescale formulation when there's only text guidance. We keep the guidance rescale hyperparameter $\phi$ to their recommended value of $0.6$.

\subsection{Evaluation}~\lblsec{appendix_eval_details}

\myparagraph{MDINOv2-I metric.} To compute this, we first detect and segment the object. For detection, we use Detic~\cite{zhou2022detecting} and Grounding DINO~\cite{liu2023grounding} in case Detic fails. For object detection, we modify the category names to be more descriptive, e.g., ``{\menlo rubber duck}'' instead of ``{\menlo toy}'',  ``{\menlo white boot}'' instead of ``{\menlo boot}'', or  ``{\menlo toy car}'' instead of ``{\menlo toy}''. We then use the detected bounding box as input to SAM~\cite{kirillov2023segment} for segmentation. Once segmented, we mask the background and crop the image around the mask for both reference and generated images. Additionally, for reference images, we manually correct the predicted mask using the SAM interactive tool to be the ground truth.

\myparagraph{Human preference study details.}
For each human preference study, we randomly sample $750$ images, with one image per object-prompt combination. We use Amazon Mechanical Turk for our study. During the study, participants first complete a practice test consisting of three questions that test their ability to select an obvious ground truth image based on alignment to the text prompt, reference object similarity, and image quality. A sample set of practice questions is shown in \reffig{practice_fig}. The study has a similar setup, except the two images are now from ours and a baseline method. We only considered responses from participants who answered the practice questions correctly.

\section{Limitations and Societal Impact}
\lblsec{limitation}
Here, we provide examples to show the limitations of our model and discuss its broader societal implications. One notable limitation of our method is that it can result in fewer variations in viewpoint and pose if the input reference images also depict the object in very similar backgrounds and poses, as shown in \reffig{limitation_fig}. Another concern is regarding the potential for synthetic data to introduce artifacts~\cite{yoon2024model}. To analyze this, \reffig{badsyncd} shows an extreme scenario of fine-tuning the model on two low-quality subsets: (1) low aesthetic score ($\leq 5$), and (2) only deformable objects. Training on low-aesthetic data preserves identity but introduces artifacts, such as floating objects (also often seen in low-quality samples due to depth guidance). Training only on deformable objects hurts identity preservation for rigid objects. This underscores the importance of our quality filtering and Objaverse category selection.
Despite these limitations, our method improves upon current leading encoder-based customization methods by proposing advancements in dataset collection, training, and inference. 

We hope this will empower users in their creative endeavors to generate ever-new compositions of concepts from their personal lives. However, the potential risks of generative models, such as creating deepfakes or misleading content, extend to our method as well. Possible ways to mitigate such risks are technologies for watermarking~\cite{fernandez2023stable} and reliable detection of generated images~\cite{wang2020cnn,corvi2022detection,cazenavette2024fakeinversion}.

\section{Change log}
\textbf{v1:} Original draft.

\noindent\textbf{v2:} Updated ICCV camera ready draft with results on FLUX~\citep{flux} model fine-tuned on our dataset.

\clearpage

\begin{figure*}[!t]
    \centering
    \includegraphics[width=0.98\linewidth]{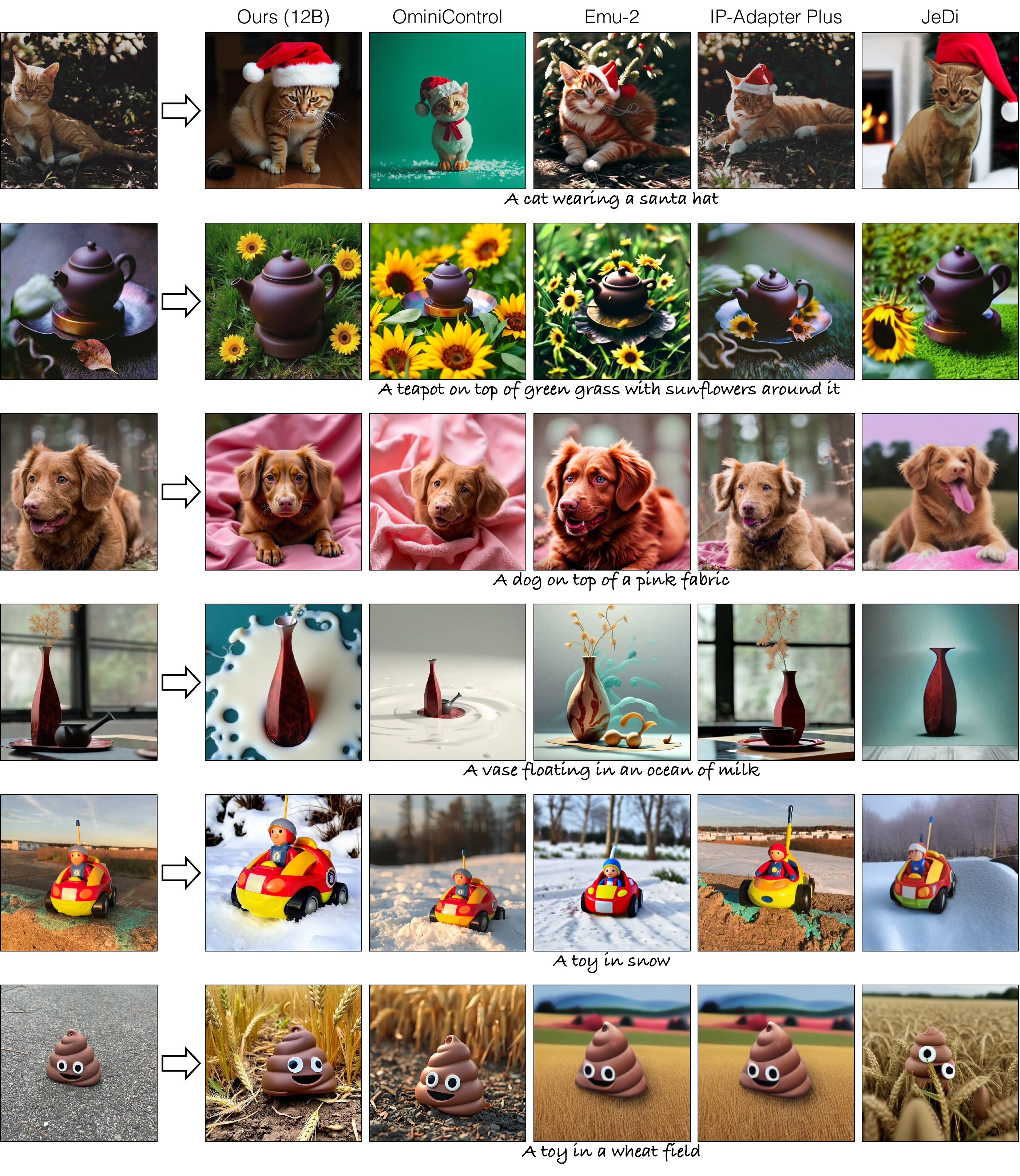}
    \vspace{-10pt}
    \caption{{\textbf{Qualitative comparison with $1$ input image.} Our method can also work with a single reference image as input, and we show a qualitative comparison here against other baselines with $1$ input image. We can successfully incorporate the text prompt while preserving the object identity similar to or higher than the baseline methods. We pick the best out of $4$ images for all methods. OminiControl can sometimes generate unrealistic images, e.g., the dog in the $3^{\text{rd}}$ row. Emu-2 and JeDi often have low fidelity, and IP-Adapter Plus overfits on the input image. \textbf{Please zoom in for details.}
    }}
    \lblfig{results_comparison_1ref_1}
    \vspace{-10pt}
\end{figure*}

\begin{figure*}[!t]
    \centering
    \includegraphics[width=0.98\linewidth]{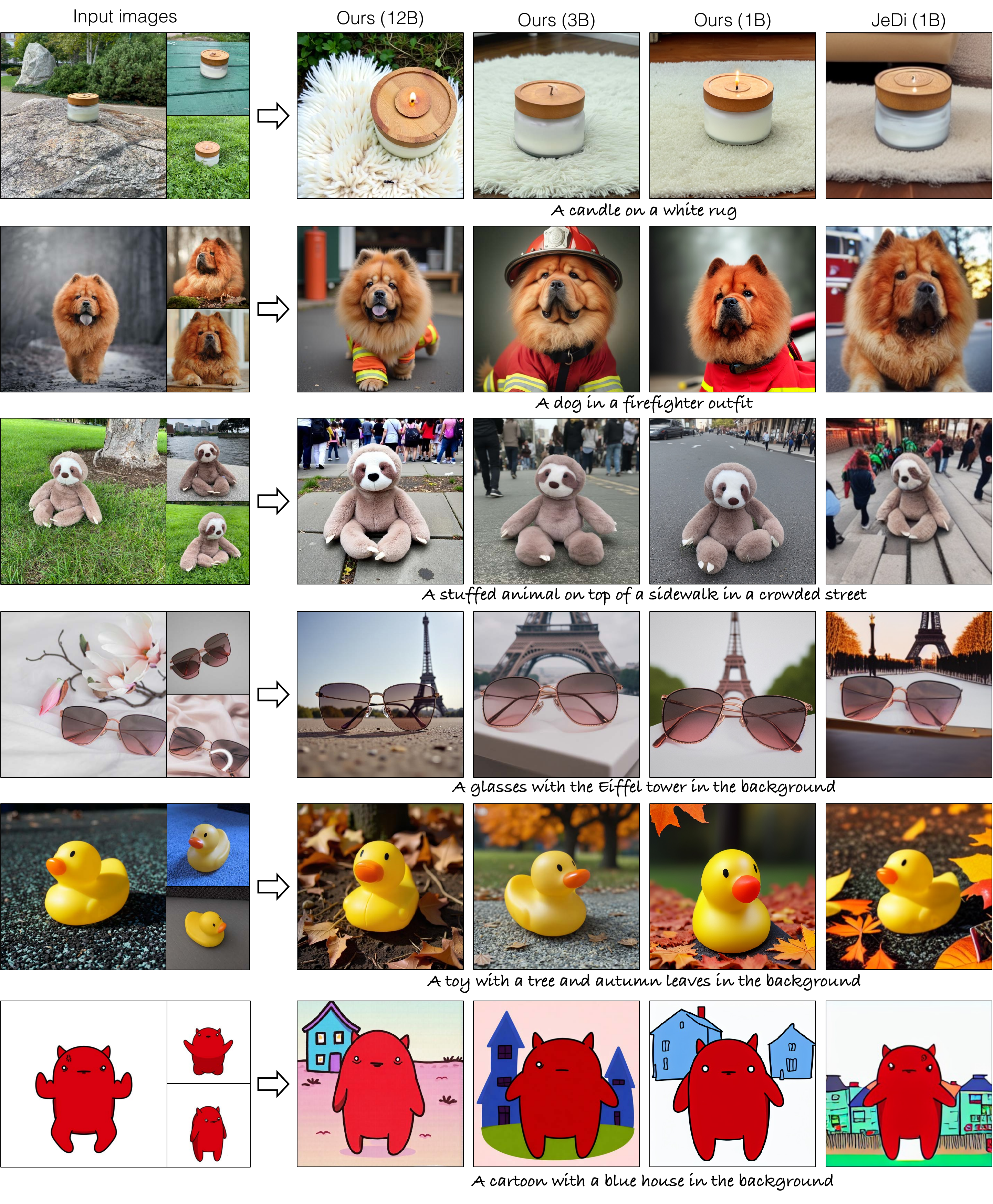}
    \vspace{-10pt}
    \caption{{\textbf{Qualitative comparison with $3$ input images.} We compare our method qualitatively against JeDi~\cite{zeng2024jedi}, which can also take multiple images as input. Compared to JeDi, our method more coherently incorporates the text prompt with higher image fidelity while being similar in performance on image alignment, e.g., the missing firefighter outfit in $2^{\text{nd}}$ row or low fidelity sunglasses in $4^{\text{rth}}$ row. We pick the best out of $4$ images for all methods. \textbf{Please zoom in for details.}
    }}
    \lblfig{results_comparison_3ref}
    \vspace{-10pt}
\end{figure*}

\begin{figure*}[!t]
    \centering
    \includegraphics[width=0.98\linewidth]{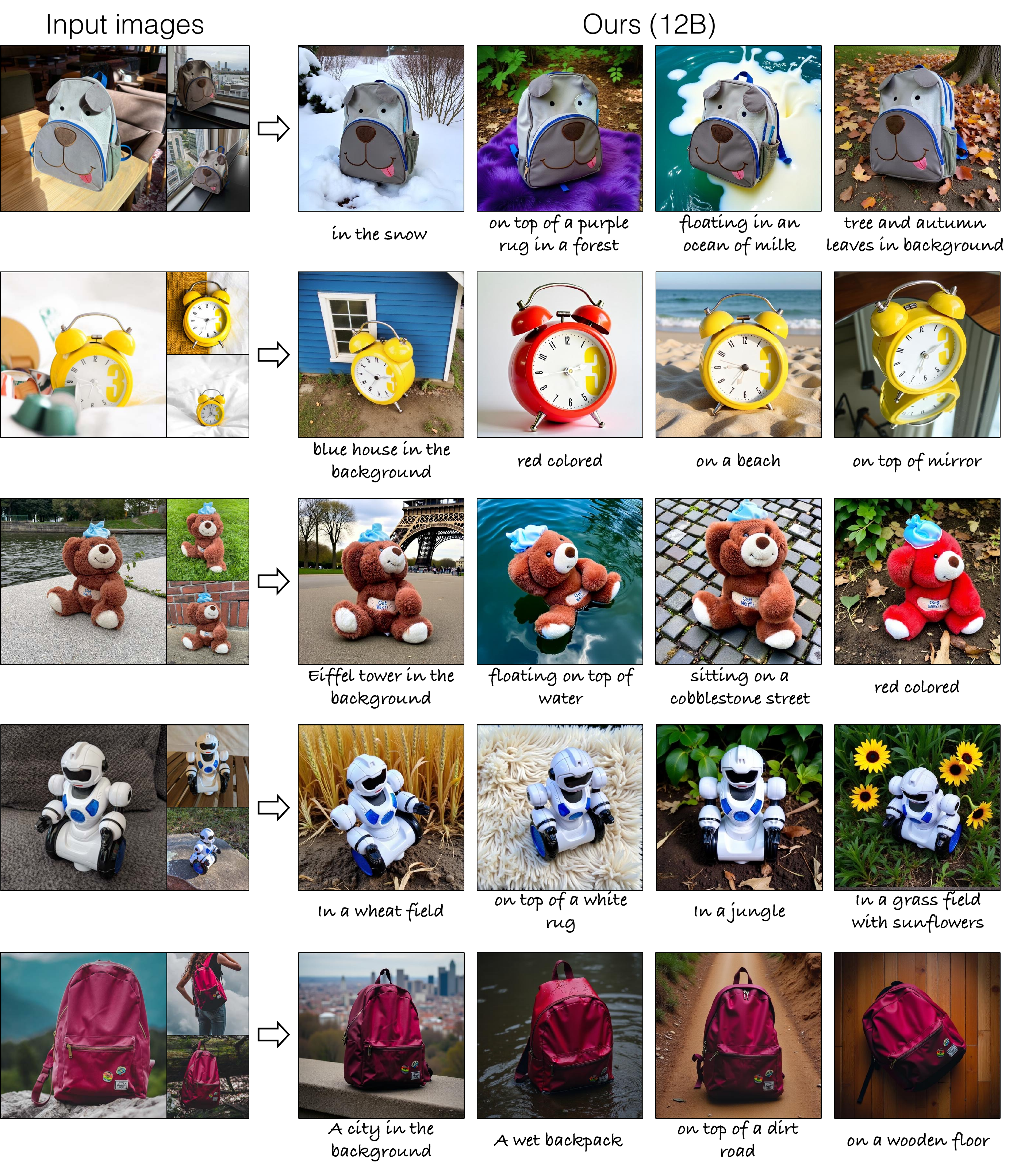}
    \vspace{-10pt}
    \caption{{\textbf{Samples on DreamBooth~\cite{ruiz2022dreambooth} dataset with $3$ input images.} We show more samples of our method given $3$ reference images of the object.
    }}
    \lblfig{random_samples_1}
    \vspace{20pt}
\end{figure*}

\begin{figure*}[!t]
    \centering
    \includegraphics[width=0.98\linewidth]{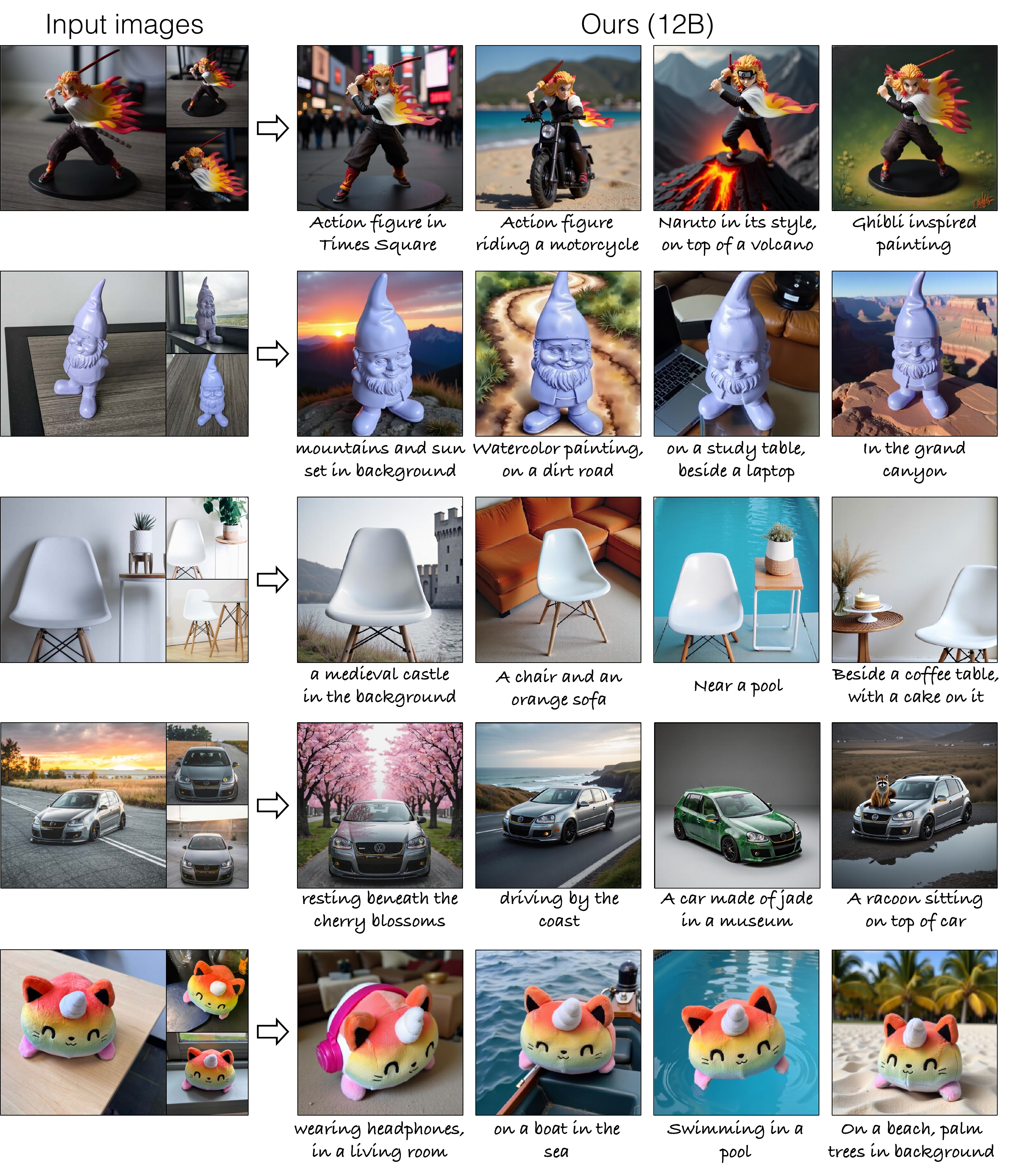}
    \vspace{-10pt}
    \caption{{\textbf{Samples on CustomConcept101~\cite{kumari2023multi} dataset with $3$ input images.} We show more samples of our method given $3$ reference images of the object.
    }}
    \lblfig{random_samples_2}
    \vspace{20pt}
\end{figure*}